\documentclass[letterpaper]{article} 
\usepackage{aaai23}  
\usepackage{times}  
\usepackage{helvet}  
\usepackage{courier}  
\usepackage[hyphens]{url}  
\usepackage{graphicx} 
\usepackage{natbib}  
\usepackage{caption} 
\usepackage{algorithm}
\usepackage{algorithmic}
\usepackage{multirow}
\usepackage{subfigure}
\usepackage{amsmath,amssymb}
\usepackage{color}
\usepackage{hyperref}

\usepackage{newfloat}
\usepackage{listings}
\DeclareCaptionStyle{ruled}{labelfont=normalfont,labelsep=colon,strut=off} 
\floatstyle{ruled}
\newfloat{listing}{tb}{lst}{}
\floatname{listing}{Listing}
\title{CF-ViT: A General Coarse-to-Fine Method for Vision Transformer}
\author{Mengzhao Chen$^1$, Mingbao Lin$^{3}$, Ke Li$^3$, Yunhang Shen$^3$,\\ Yongjian Wu$^3$, Fei Chao$^{1}$, Rongrong Ji$^{1,2}$\thanks{Corresponding Author}}
\affiliations{$^1$MAC Lab, Department of Artificial Intelligence, Xiamen University \\
$^2$Institute of Artificial Intelligence, Xiamen University
\quad
$^3$Tencent Youtu Lab \\
{\tt\small cmzxmu@stu.xmu.edu.cn, linmb001@outlook.com, tristanli.sh.gmail.com,}\\ 
{\tt\small shenyunhang01@gmail.com, littlekenwu@tencent.com, \{fchao, rrji\}@xmu.edu.cn}}

\usepackage{bibentry}

\begin{document}

\maketitle

\begin{abstract}
Vision Transformers (ViT) have made many breakthroughs in computer vision tasks. However, considerable redundancy arises in the spatial dimension of an input image, leading to massive computational costs.
Therefore, We propose a coarse-to-fine vision transformer (CF-ViT) to relieve computational burden while retaining performance in this paper. Our proposed CF-ViT is motivated by two important observations in modern ViT models: (1) The coarse-grained patch splitting can locate informative regions of an input image. (2) Most images can be well recognized by a ViT model in a small-length token sequence. 
Therefore, our CF-ViT implements network inference in a two-stage manner. At coarse inference stage, an input image is split into a small-length patch sequence for a computationally economical classification. If not well recognized, the informative patches are identified and further re-split in a fine-grained granularity. 
Extensive experiments demonstrate the efficacy of our CF-ViT. For example, without any compromise on performance, CF-ViT reduces 53\% FLOPs of LV-ViT, and also achieves 2.01$\times$ throughput.
Code of this project is at \url{https://github.com/ChenMnZ/CF-ViT}.
\end{abstract}

\section{Introduction}\label{introduction}
Tremendous successes of traditional transformer~\cite{vaswani2017attention} in natural language processing (NLP) have inspired the researchers to go further on computer vision~\cite{han2022survey}. Consequently, vision transformers (ViT)~\cite{dosovitskiy2020image} receive ever-increasing attentions in many vision tasks such as image classification~\cite{dosovitskiy2020image,jiang2021all}, object detection~\cite{liu2021swin,wang2021pyramid}, semantic segmentation~\cite{zheng2021rethinking,xie2021segformer}, \emph{etc}.

\begin{figure}[!t]
\centering
\subfigure[]{
\includegraphics[width=0.48\linewidth]{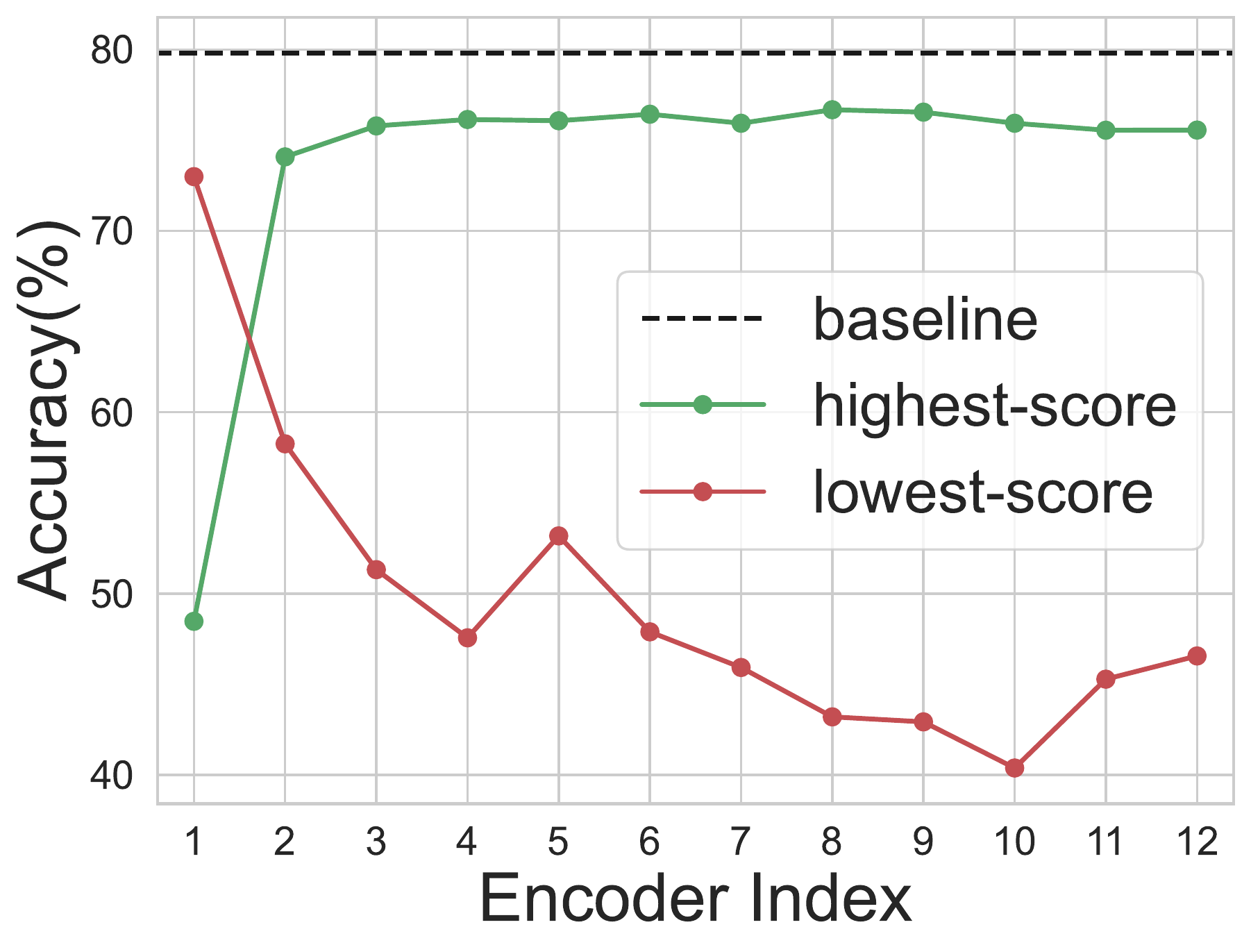}
}
\subfigure[]{
\includegraphics[width=0.48\linewidth]{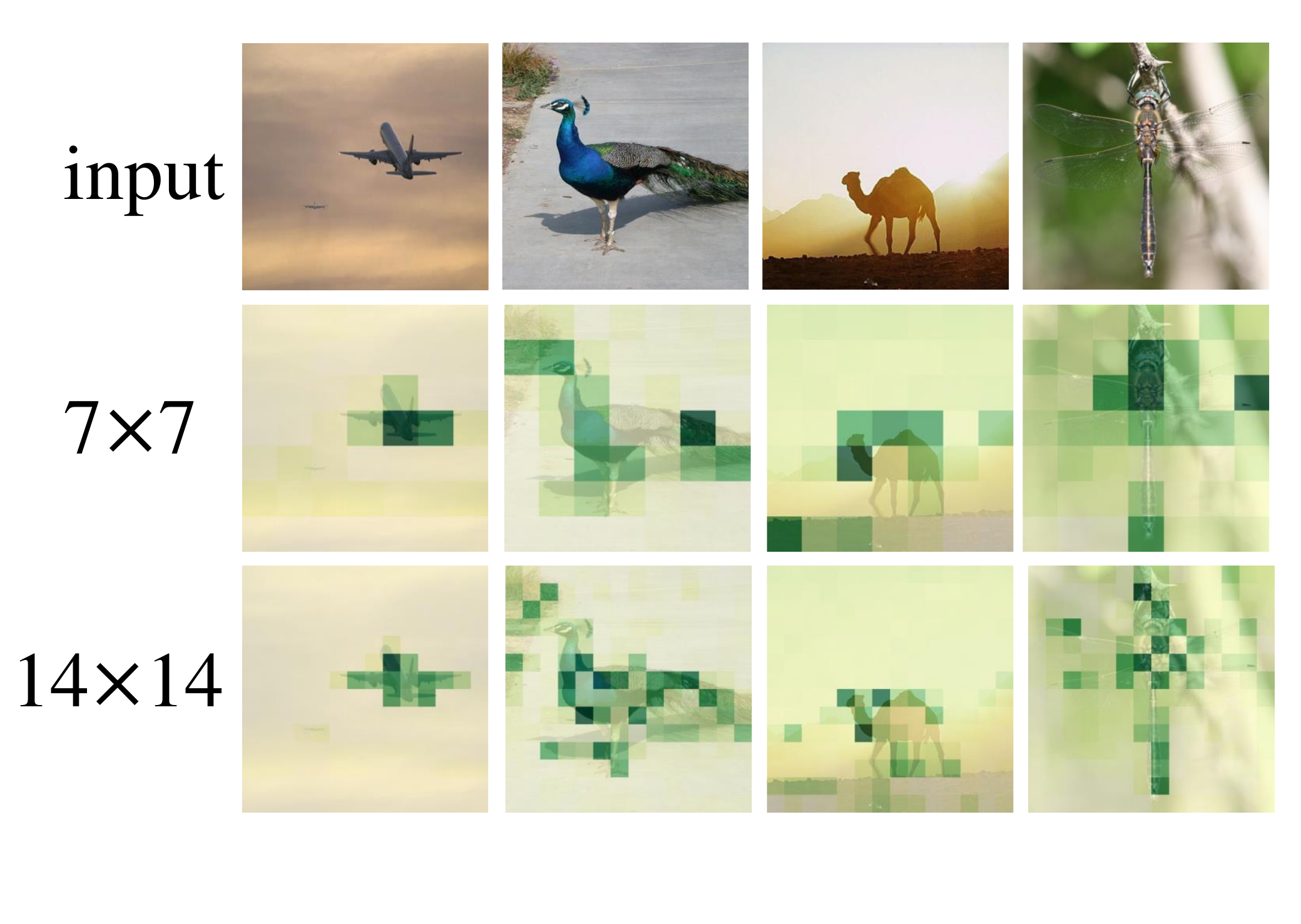}
}
\caption{\label{class_attention_fig} (a) Performance comparison (DeiT-S~\cite{touvron2021training}) between 100 highest-score patches and 100 lowest-score patches. The score is measured by the class attention. (b) Visualization of class attention in the last encoder of DeiT-S (Best viewed with zooming in).}
\end{figure}

By splitting a 2D image into a patch sequence and using a linear projection to embed these patches into 1D tokens as inputs, ViT merits in its property of modeling long-range dependencies among tokens. Generally speaking, the performance of a ViT model is closely correlated with the token number~\cite{dosovitskiy2020image,wang2021not}, to which, however, the computational cost of ViT is also quadratically related.
Fortunately, images often have more spatial redundancy than languages\,\cite{wang2014exploiting}, such as regions with task-unrelated objects. 
Thus, many works~\cite{wang2020glance,yang2020resolution,wang2022adafocus,wang2021adaptive,wang2022adafocusv3,han2022latency,han2021spatially} o try to adaptively reduce the input resolution of convolution neural networks.
Also, great efforts have been made to excavate redundant tokens for ViTs.
For example, PS-ViT~\cite{tang2022patch} enhances transformer efficiency by a top-down token pruning paradigm. Both DynamicViT~\cite{rao2021dynamicvit} and IA-RED$^2$~\cite{pan2021ia} devise a lightweight prediction module to estimate the importance score of each token, and discard low-score tokens. Following DynamicViT~\cite{rao2021dynamicvit}, EViT~\cite{liang2022evit} regards class attention as a metric of token importance, which avoids the introduction of extra parameters. Unlike discarding tokens directly, DGE~\cite{song2021dynamic} introduces sparse queries to reduce the output token number. Evo-ViT~\cite{xu2022evovit} maintains the spatial structure while consuming less computational cost to update uninformative tokens. DVT~\cite{wang2021not} cascades multiple ViTs with increasing tokens, then leverages an early-exiting policy to decide each image's token number.

This paper observes two insightful phenomena.
First, similar to the fine-grained patch splitting like 14$\times$14, the coarse-grained patch splitting such as 7$\times$7 can also locate the informative regions. To verify this, we first show that class attention \big(see Eq.\,(\ref{attention_equal})\big)~\cite{liang2022evit} has a remarkable ability to identify informative patches. We conduct a toy experiment on the validation set of ImageNet~\cite{deng2009imagenet} with a pre-trained DeiT-S model~\cite{touvron2021training}. Following DeiT-S, we split each image into 14$\times$14 patches, then compute the class attention score of each patch within each encoder. Then, the 100 highest-score patches and 100 lowest-score patches are respectively fed to DeiT-S to obtain their accuracy in Fig.\,\ref{class_attention_fig}(a). In the figure, highest-score patches outperform lowest-score ones by a large margin, such a result demonstrates that class attention can well reflect more informative patches. Then, in Fig.\,\ref{class_attention_fig}(b), we visualize the class attention in the last encoder of DeiT-S. As can be seen, 7$\times$7 splitting and 14$\times$14 splitting generally obtain similar attentive regions.

\begin{table}[t]
\setlength{\tabcolsep}{2pt}
\centering
\caption{\label{different_token_acc}Accuracy and FLOPs of Deit-S~\cite{touvron2021training} on ImageNet with different No. of patches as inputs.}
\label{table:headings}
\begin{tabular}{c|ccc}
\hline\noalign{\smallskip}
No. of token & 14$\times$14 & 7$\times$7 \\
\hline
Accuracy & 79.8\% & 73.2\% \\
FLOPs & 4.60G & 1.10G \\
\hline
\end{tabular}   
\end{table}

The second phenomenon is that most images can be well recognized by a ViT model in a small-length patch sequence. We train Deit-S~\cite{touvron2021training} with varying lengths of input sequence, and report top-1 accuracy and FLOPs in Table\,\ref{different_token_acc}, in which, with 4.2$\times$ higher computational cost, splitting a 2D image into fine-grained 14$\times$14 patches only obtains 6.6\% accuracy benefits than coarse-grained 7$\times$7 splitting. A similar 
observation has been discussed in DVT~\cite{wang2021not}. This indicates that most regions in 73.2\% images are ``easy'' so that coarse-grained 7$\times$7 patch splitting can well implement the classification, and a small portion of images are filled with ``hard'' regions requiring a fine-grained splitting of 14$\times$14 with heavier computation. 
%
%
Thus, we split the ``easy'' samples with coarse-grained patch splitting for cheaper computation. In addition, for ``hard'' samples, we can differentiate ``easy'' regions and ``hard'' regions, and split them with different patch sizes in order to pursue an efficient inference while retaining good performance.  
Note that, these two phenomena can be found in other ViT models as well.

\begin{figure}[t]{}
\centering
\includegraphics[width=0.9\linewidth]{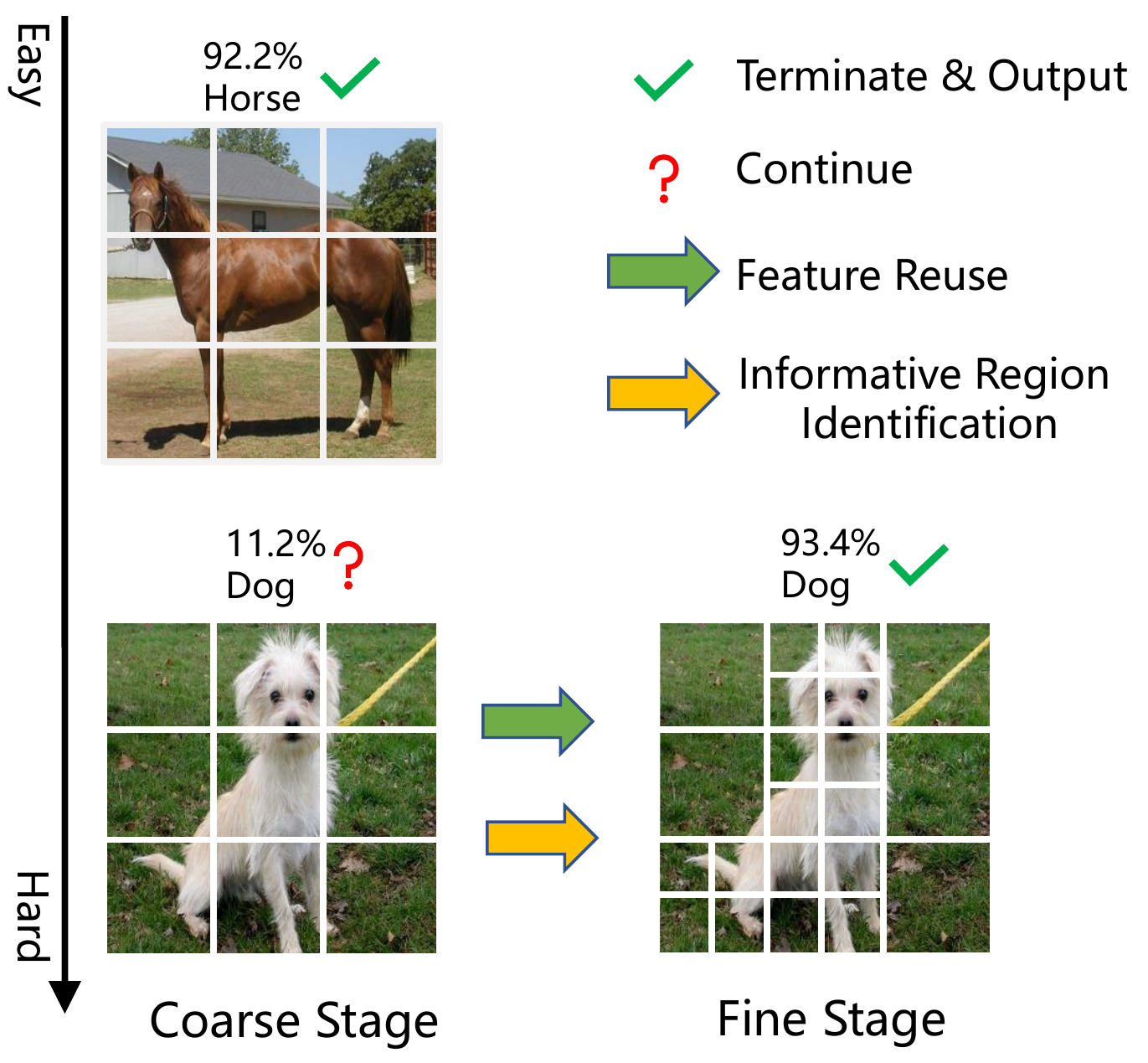}
\caption{\label{example_fig}Example of CF-ViT.}
\end{figure}

Inspired by the above observations, we propose a novel coarse-to-fine vision transformer in this paper, termed CF-ViT, which aims to produce correct predictions with input-adaptive computational cost. As shown in Fig.\,\ref{example_fig}, the inference of CF-ViT is divided into a coarse inference stage and a fine inference stage. The coarse stage receives coarse-grained patches as network inputs, which merits in low computational cost since the coarse-grained splitting results in much fewer patches (tokens). If this stage owns a high confidence score, the network inference terminates; otherwise, the input image is split again into fine-grained patches and fed to the fine stage. In contrast to a naive re-splitting of all coarse patches, we design a mechanism of informative region identification to further split the informative patches in a fine-grained state, and retain these patches with less information in a coarse-grained state. Therefore, we avoid heavy computational burdens on the redundant image patches. Also, we introduce a mechanism of feature reuse to inject the integral information of the coarse patch to the split fine-grained patches, which further enhance the model performance. We then evaluate our proposed CF-ViT built upon DeiT~\cite{touvron2021training} and LV-ViT~\cite{jiang2021all} on ImageNet~\cite{deng2009imagenet}. Extensive experiments results show that our CF-ViT can well boost the inference efficiency. For example, without any performance compromise, CF-ViT reduces 53\% FLOPs of LV-ViT and also leads to $2.01\times$ practical throughput on an A100 GPU. In addition, extensive ablation studies also demonstrate the efficacy of each design in our CF-ViT including the informative region identification and the feature reuse.
Visualization of inference results shows that CF-ViT enables adaptive inference according to the ``difficulty'' of input images and can accurately locate informative regions for re-splitting in fine inference stage.

\section{Related Work}
\subsection{Vision Transformer}
Motivated by successes of transformer~\cite{vaswani2017attention} in NLP, researchers develop vision transformer (ViT)~\cite{dosovitskiy2020image} for image recognition. 
However, the lack of inductive bias~\cite{dosovitskiy2020image} requires ViT model to be pre-trained on a very large-scale data corpus such as JFT-300M~\cite{sun2017revisiting} to pursue a desired performance. This demand barricades the development of ViT model since the large-scale dataset requires a high-capacity workstation. To handle this, many studies~\cite{touvron2021training,jiang2021all} develop specialized training strategies for ViT models. For instance, DeiT~\cite{touvron2021training} introduces an extra token for knowledge distillation. LV-ViT~\cite{jiang2021all} leverages all tokens to compute the training loss, and the location-specific supervision label of each patch token is generated by a machine annotator. In addition, another group focuses on improving the architecture of ViT~\cite{yuan2021incorporating,li2021localvit,chu2021conditional,chu2021twins,yuan2021tokens,heo2021rethinking,touvron2021going,liu2021swin,li2022efficient}. For example, CPVT~\cite{chu2021conditional} uses a convolution layer to replace the learnable positional embedding. CaiT~\cite{touvron2021going} builds deeper transformers and develops specialized optimization strategies for training.
TNT~\cite{han2021transformer} leverages an inner block to model the pixel-wise interactions within each patch, which merits in preserving more rich local features. Our study in this paper introduces a general framework, which aims to improve the inference efficiency of various ViT backbones in an input-adaptive manner.

\subsection{ViT Compression}
Except to develop high-performing ViT models, the expensive computation of a ViT model also arouses wide attention and methods are explored to facilitate ViT deployment. Based on whether the inference is input-dependent, we empirically categorize existing studies on compressing ViTs into two groups below.

\textbf{Static ViT Compression}. This group mainly focuses on reducing the network complexity through manually designed modules with a fixed computational graph regardless of the input images. Inspired by the great success of hierarchical convolutional neural networks in dense prediction tasks such as segmentation and detection, recent advances~\cite{heo2021rethinking,liu2021swin,pan2021scalable,yuan2021tokens,wang2021pyramid} introduce hierarchical transformers. Also, many others~\cite{liu2021swin,huang2021shuffle,fang2021msg,yu2021glance} consider local self-attention to reduce the complexity of traditional global self-attention.
Compared to these static methods using a fixed computational graph, our CF-ViT, improves the inference efficiency by adaptively selecting an appropriate computational path for each image.

\textbf{Dynamic ViT Compression}.
In contrast to static ViT compression, dynamic ViT adapts the computational graph according to its input images~\cite{han2021dynamic}. Some works~\cite{rao2021dynamicvit,pan2021ia,liang2022evit,lin2022super} attempt to dynamically prune these tokens considered unimportant during inference. On the contrary, Evo-ViT~\cite{xu2022evovit} chooses to preserve the unimportant tokens, which however, are assigned with a lower computational budget for updating. Unlike these pruning-based methods, our CF-ViT maintains the integrity of image information and reinforces the inference efficiency by enlarging the size of patches in uninformative regions, which leads to fewer tokens. The rationale behind this is that uninformative regions like background contribute less to the recognition thus a fine-grained patch splitting is unnecessary. Note that, the recent DVT~\cite{wang2021not} endows a proper token number for each input image by cascading three transformers. Though meriting in inference acceleration, it inevitably increases the storage overhead by 3$\times$. Different from DVT, our CF-ViT trains only one transformer that can accept different sizes of input tokens, and only conducts fine-grained token splitting on informative regions rather than an entire image. QuadTree~\cite{tang2022quadtree} also builds token pyramids in a coarse-to-fine manner. However, QuadTree performs coarse-to-fine splitting for all images while our CF-ViT only performs fine splitting on “hard” images to further reduce computation cost.

\section{Preliminaries}\label{preliminaries}
Vision Transformer(ViT)~\cite{dosovitskiy2020image} splits a 2D image into flattened 2D patches and uses an linear projection to map patches into tokens, \emph{a.k.a}. patch embeddings. Besides, an extra [class] token, which represents the global image information, is appended as well. Moreover, all tokens are added with a learnable positional embedding. Thus, the input token sequence of a ViT model is:
\begin{align}\label{embedding_equal}
\mathbf{X}_0 = [\mathbf{x}_0^0;\mathbf{x}_0^1;...;\mathbf{x}_0^N] + \mathbf{E}_{pos} ,
\end{align}
where $\mathbf{x}_0^i \in \mathbb{R}^{D}$ is a $D$-dimensional token of the $i$-th patch if $i > 0$, and [class] token if $i = 0$. The $\mathbf{E}_{pos}$ and $N$ are the position embedding and patch number.

A ViT model $\mathcal{V}$ contains $K$ sequentially stacked encoders, each of which consists of a self-attention (SA) module\footnote{SA has been replaced by multi-head self-attention (MHSA) in most ViTs. For brevity, we simply discuss SA herein. Also, we ignore the shortcut operation in Eq.\,(\ref{ffn_equal}).} and a feed-forward network (FFN). In SA of the $k$-th encoder, the token sequence $\mathbf{X}_{k-1}$ is projected into a query matrix $\mathbf{Q}_{k} \in \mathbb{R}^{(N+1) \times D}$, a key matrix $\mathbf{K}_{k} \in \mathbb{R}^{(N+1) \times D}$, and a value matrix $\mathbf{V}_{k} \in \mathbb{R}^{(N+1) \times D}$. Then, the self-attention matrix $\mathbf{A}_{k}  \in \mathbb{R}^{(N+1) \times (N+1)}$ is computed as:
\begin{align}\label{attention_equal}
\mathbf{A}_k = \mathrm{Softmax}(\frac{\mathbf{Q}_k \mathbf{K}_k^{T}}{\sqrt{D}}) = [\mathbf{a}_k^0;\mathbf{a}_k^1;...;\mathbf{a}_k^N].
\end{align}

The $\mathbf{a}_k^0 \in \mathbb{R}^{(N+1)}$ is known as class attention, reflecting the interactions between [class] token and other patch tokens.
With $\mathbf{A}_k$, the outputs of SA, \emph{i.e.}, $\mathbf{A}_k\mathbf{V}_k$, are sent to FFN consisting of two fully-connected layers to derive the updated tokens $\mathbf{X}_k = [\mathbf{x}_k^0; \mathbf{x}_k^1;...;\mathbf{x}_k^N]$.
The [class] token $\mathbf{x}^0_k$ is derived as:
\begin{equation}\label{ffn_equal}
    \mathbf{x}^0_k = FFN(\mathbf{a}^0_k\mathbf{V}_k).
\end{equation}

After a series of SA-FFN transformations, the [class] token $\mathbf{x}^0_K$ from the $K$-th encoder is fed to the classifier to predict the category of the input.

\textbf{ViT Complexity}. Given that an image is split into $N$ patches, the computational complexity of SA and FFN are~\cite{xu2022evovit}:
\begin{equation}\label{complexity_equal}
\begin{split}
O(S&A) = 3ND^2 + 2 N^2 D, \\&
O(FFN) = 8ND^2.
\end{split}
\end{equation}

We can see that the complexities of SA and FNN are respectively quadratic and linear to $N$. Thus, ViT complexity can be well reduced by decreasing the input patch (token) number~\cite{rao2021dynamicvit,pan2021ia,liang2022evit,wang2021not}, which is also the focus of this paper.

\begin{figure*}[!ht]
\centering
\includegraphics[width=0.8\textwidth]{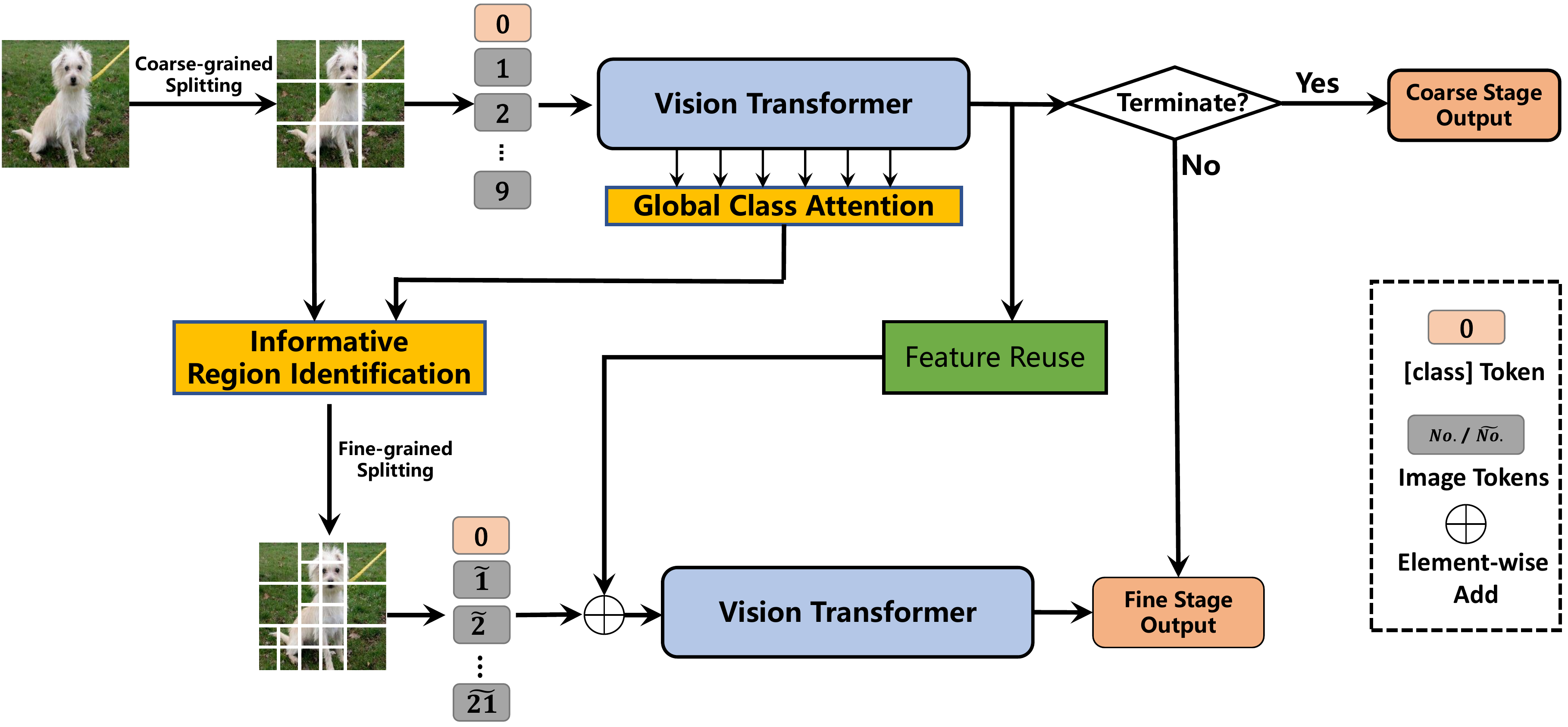}
\caption{\label{framework_fig}Framework of our CF-ViT. Note that both coarse stage and fine stage share the same network parameters. }
\end{figure*}

\section{Coarse-to-Fine Vision Transformer}

This section formally introduces our CF-ViT that decreases the computational cost by reducing the input sequence length. Our motive lies in two observations in Sec.\,\ref{introduction}: (1) The coarse-grained patch splitting can well locate the informative objects as well. (2) Most images can be well recognized by a ViT model in a small sequence length. These inspire us to implement a ViT in a two-stage manner. As shown in Fig.\,\ref{framework_fig}, the coarse inference stage implements image recognition with a small length of token sequence. If not well recognized, the informative regions will be further split for a fine-grained recognition. Details are given below.

%

\subsection{Coarse Inference Stage}\label{coarse_inference_stage}
CF-ViT first performs a coarse splitting to recognize images filled with ``easy'' regions. Also, it locates informative regions for an efficient inference when meeting ``hard'' samples. At coarse stage, the input of our CF-ViT model $\mathcal{V}$ is:
\begin{align}\label{coarse_embedding_equal}
\mathbf{X}_0^{c} = [\mathbf{x}_{0}^0;\mathbf{x}_0^1;...;\mathbf{x}_0^{N_{c}}] + \mathbf{E}_{pos}^{c},
\end{align}
where $N_{c}$ is the number of coarse patches. Supposing $\mathcal{V}$ contains $K$ encoders, after SA-FFN transformations in Sec.\,\ref{preliminaries}, the output token sequence of $\mathcal{V}$ is:
\begin{equation}\label{coarse_stage_inferece}
\begin{split}
\mathcal{V}(\mathbf{X}_0^{c})  =[\mathbf{x}_K^0;\mathbf{x}_K^1;...;\mathbf{x}_K^{N_{c}}].
\end{split}
\end{equation}

Finally, the [class] token $\mathbf{x}_{K}^0$ is fed to a classifier $\mathcal{F}$ to obtain the coarse-stage category prediction distribution $\mathbf{p}^{c}$:
\begin{equation}
    \mathbf{p}^{c} = \mathcal{F}(\mathbf{x}_{K}^0)=[p^c_1, p^c_2, ..., p^c_n],
\end{equation}
where $n$ denotes the  category number. So far, we can obtain the predicted category of the input as:
\begin{equation}
    j = \mathop{\arg\max}\limits_{i} \; p^c_i.
\end{equation}

We expect a large $p^c_j$ since it serves as a prediction confidence score at coarse inference stage where the computational cost is very cheap due to a small value of patch number $N_c$. We introduce a threshold $\eta$ to realize a trade-off between performance and computation. In our implementation, if $p^c_j \ge \eta$, the inference will terminate and we attribute the input to category $j$. Otherwise, the input images might contain ``hard'' regions undistinguished to the ViT model. Thus, a more fine-grained patch splitting is urgent.

\textbf{Informative Region Identification.} 
The most naive solution is to further split all the coarse patches $[\mathbf{x}_0^1;\mathbf{x}_0^2;...;\mathbf{x}_0^{N_{c}}]$. However, the drastically increasing tokens inevitably cause severe computational costs. Instead, for an economical budget, we propose to identify and then re-split these informative regions that are the most beneficial to the performance increase. Thus, the key now lies in how to identify the informative patches. 

Recall that, the class attention $\mathbf{a}^0_k \in \mathbf{A}_k$ in Eq.\,(\ref{attention_equal}) reflects the interactions between [class] token and other image patch tokens in the $k$-th encoder. Besides, the [class] token $\mathbf{x}_k^0 = FFN(\mathbf{a}_k^0\mathbf{V}_k)$, which indicates that each item $(\mathbf{a}_k^0)_i$ models the weighted coefficient of the $i$-th token $\mathbf{x}^i_0$ to the performance. Therefore, it is natural to use the class attention $\mathbf{a}^0_k$ as a score to indicate if a token is informative. From Fig.\,\ref{class_attention_fig}(a), we can see that, the performance of preserving 100 highest-score patches usually outperforms that of 100 lowest-score patches. This demonstrates that class attention can be a reliable measure. Nevertheless, we can also observe the instability of class attention in the bottom layers, such as the first encoder. To overcome it, we propose global class attention  that combines class attention across different encoders using exponential moving average (EMA) to better identify informative patches:
\begin{align}\label{global_class_attention_equal}
\bar{\mathbf{a}}_{k} = \beta \cdot \bar{\mathbf{a}}_{k-1} + (1-\beta) \cdot \mathbf{a}_{k}^{0}, 
\end{align}
where $\beta =0.99$. The global class attention begins from the 4-th encoder and we select patches with high-score global class attention in the last encoder $\bar{\mathbf{a}}_K$.

Earlier studies~\cite{liang2022evit,xu2022evovit} also adopt class attention to indicate token importance. This paper differs in two folds: (1) A comprehensive demonstration on high-score patches are given in Fig.\,\ref{class_attention_fig}(a). (2) We consider the global class attention instead of class attention in a particular layer~\cite{liang2022evit}, efficacy of which is given in Table.\,\ref{selection_ablation_table}.

\begin{figure*}[!t]
\centering
\includegraphics[width=0.8\textwidth]{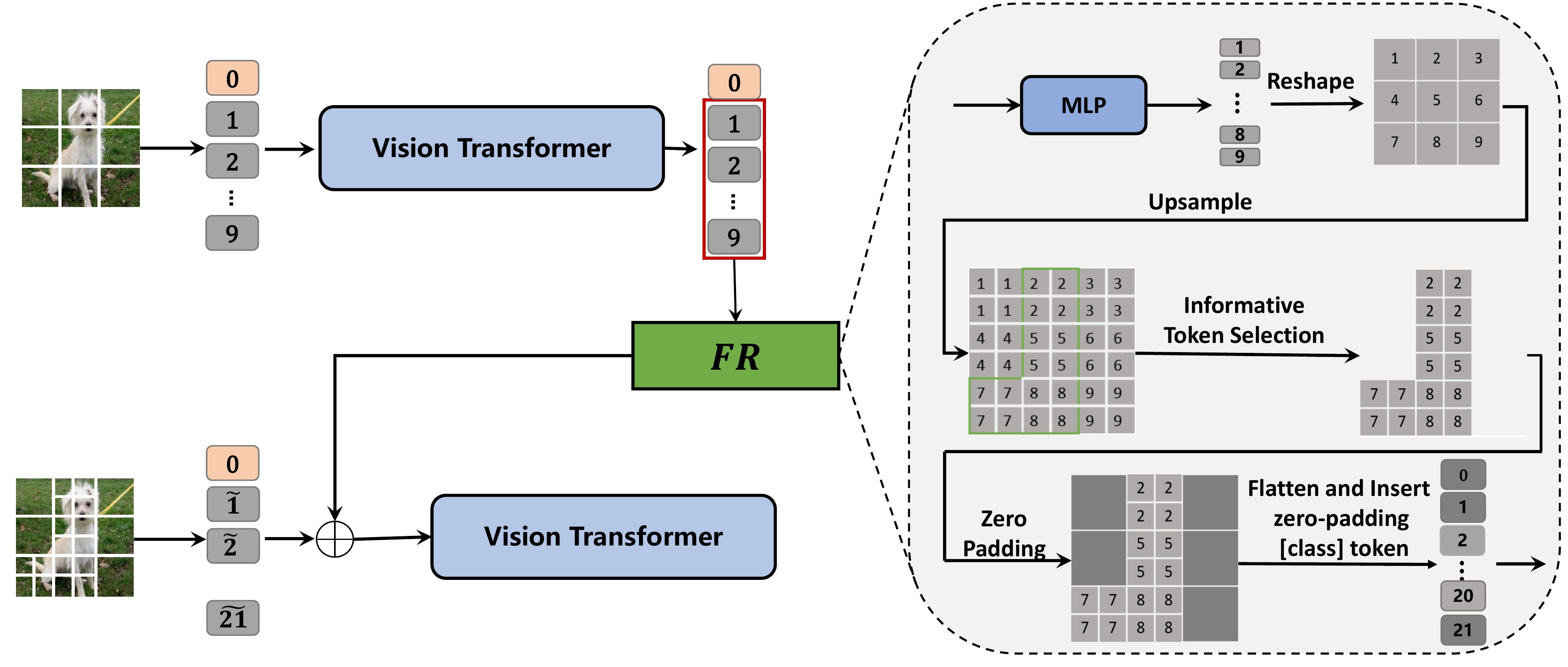}
\caption{\label{reuse_fig}Illustration of our feature reuse module. An MLP is firstly introduced for a flexible transformation among token sequences from coarse stage. Then, four copies of each token are executed as an upsampling strategy and these areas corresponding to fine-grained splitting patches are selected to shortcut to the fine-grained token sequence.}
\end{figure*}

\subsection{Fine Inference Stage}\label{fine_inference_stage}
With the global class attention $\bar{\mathbf{a}}_K$ on hand, we continue to perform a fine-grained splitting on informative patches when the prediction in coarse stage $p_j^c < \eta$, which indicates an indistinguishable input image.

To that effect, we pick up these coarse patches, attention scores of which are within the top-$\alpha N_c$ largest among the $N_c$ coarse patches, where $\alpha \in [0, 1]$ represents the rate of informative patches. Then, as shown in Fig.\,\ref{framework_fig}, each informative patch is further split into $2 \times 2$ patches for better representation in a finer granularity. Consequently, the patch number after fine-grained splitting is:
\begin{align}\label{input_num_equal}
N_{f} = 4 \lceil N_{c}\alpha \rceil + \lfloor N_{c}(1-\alpha) \rfloor,
\end{align}
where $\lceil \cdot \rceil$ and $\lfloor \cdot \rfloor$ respectively round up and round down their inputs. It is intuitive that $\alpha$ provides a trade-off between accuracy and efficiency. The $\alpha = 0$ indicates no fine inference and results in the fewest patches. Though computationally economical, performance drops if the test set is full of ``hard'' images. In contrast, $\alpha = 1$ leads the fine inference stage of our CF-ViT degenerates to traditional ViT models~\cite{jiang2021all,touvron2021training}. In this case, the computational cost is extraordinarily expensive. For a trade-off, we set $\alpha$ to 0.5 in our implementation.

\textbf{Feature Reuse.} 
After re-splitting of each input image, the input token sequence at fine inference stage of our CF-ViT becomes:
\begin{equation}\label{fine_input}
    \tilde{\mathbf{X}}_0^f = [\mathbf{x}_0^0;\tilde{\mathbf{x}}_0^1;...;\tilde{\mathbf{x}}_0^{N_f}] + \mathbf{E}_{pos}^f.
\end{equation}

Suppose that an informative patch $\mathbf{x}_0^i$ in Eq.\,(\ref{coarse_embedding_equal}) is further split into $2\times2$ patches in Eq.\,(\ref{fine_input}), which offer a finer granularity. Nevertheless, they also cut off the integrity of the local patch $\mathbf{x}_0^i$. To solve this, we also devise a feature reuse module to inject the information of $\mathbf{x}_0^i$ into the four fine-grained patches.

Fig.\,\ref{reuse_fig} illustrates our feature reuse. It takes the output token sequence from coarse stage as input. Similar to FNN, the input will be processed by an MLP layer first to allow a flexible transformation. Then, these transformed tokens are reshaped and each of them is copied 4$\times$. Further, these tokens corresponding to fine-grained splitting patches are picked up as the output of feature reuse module, denoted as $\mathbf{X}_{r} = FR([\mathbf{x}_K^1;\mathbf{x}_K^2;...;\mathbf{x}_K^{N_{c}}])$. We do not consider reusing [class] token and uninformative tokens by zeroing out them, since we empirically find them unbeneficial to the performance as verified in Table.\,\ref{reuse_ablation_table}. Finally, we shortcut $\mathbf{X}_r$ to the re-split token sequence $\tilde{\mathbf{X}}_0^f$ in Eq.\,(\ref{fine_input}) as the final input of our CF-ViT model $\mathcal{V}$ at fine inference stage. Consequently, the output of $\mathcal{V}$ is:
\begin{equation}\label{fine_stage_inferece}
\begin{split}
\mathcal{V}(\tilde{\mathbf{X}}_0^{f} + \mathbf{X}_r)  =
[\tilde{\mathbf{x}}_K^0;\tilde{\mathbf{x}}_K^1;...;\tilde{\mathbf{x}}_K^{N_{f}}].
\end{split}
\end{equation}

Finally, the [class] token $\tilde{\mathbf{x}}_{K}^0$ is fed to the same classifier $\mathcal{F}$ to obtain the fine-stage category prediction distribution $\mathbf{p}^{f}$:
\begin{equation}
    \mathbf{p}^{f} = \mathcal{F}(\mathbf{x}_{K}^0)=[p^f_1, p^f_2, ..., p^f_n].
\end{equation}

\subsection{Training Strategy}\label{training_strategies}
During the training of our CV-ViT, we set the confidence threshold $\eta = 1$, which means the fine inference stage will be always executed for every input image. On the one hand, we expect the fine-grained splitting can well fit the ground truth label $\mathbf{y}$ for an accurate prediction of the input. On the other hand, we expect the coarse-grained splitting obsesses a similar output with that of fine-grained splitting such that most input can be well recognized at coarse inference stage, which indicates less computational cost. Consequently, the training loss of our CF-ViT is given below:
\begin{equation}\label{loss_equation}
    loss = CE(\mathbf{p}_f, \mathbf{y}) + KL(\mathbf{p}_c, \mathbf{p}_f),
\end{equation}
where $CE(\cdot,\cdot)$ and $KL(\cdot,\cdot)$ respectively represent the cross entropy loss and Kullback-Leibler divergence.

During the inference of our CV-ViT, by varying the value of $\eta$, we can obtain a trade-off between computational budget and accuracy performance. A large $\eta$ means more inputs will be sent to fine inference stage, which indicates better performance but more computational cost, and vice versa.

\section{Experiments}

\subsection{Implementation Details}


%
For ease of comparison, following existing studies on ViT~\cite{rao2021dynamicvit,liang2022evit,tang2022patch,xu2022evovit}, we build our CF-ViT with DeiT-S (w/o distillation)~\cite{touvron2021training} and LV-ViT-S~\cite{jiang2021all} as backbone networks, all of which split each image into $14 \times 14 = 196$ patches. 
To show the advantages of our CF-ViT, we conduct the experiments on ImageNet~\cite{deng2009imagenet} from two perspectives: 
(1) Each input image is split into $7\times7$ ($N_c = 49$) patches at coarse inference stage, leading to a total of $N_f = 124$ patches at fine inference stage according to Eq.\,(\ref{input_num_equal}). As results, the computation of our CF-ViT is much cheaper than its backbones due to its less split patches.
(2) The image patches number at coarse inference stage is changed to $9 \times 9$, leading to 204 patches at fine inference stage, which maintains similar FLOPs with the backbones. For a clearer presentation, CF-ViT is denoted as CF-ViT$^{*}$ in this case.

\begin{table}[t]
\centering
\setlength{\tabcolsep}{3pt}
\caption{\label{throughput_table} Comparison between CF-ViT and its backbones. }
\begin{tabular}{ccccc}
\hline\noalign{\smallskip}
\multirow{2}{*}{Model} & \multirow{2}{*}{$\eta$} & Top-1 Acc. & FLOPs & Throughput\\
 & & (\%) & (G) &(img./s) \\
\hline
DeiT-S & - & 79.8 & 4.6 & 2601 \\
\hline
CF-ViT & 0.5 & 79.8{\color{blue}($+0.0$)} & 1.8{\color{blue}($\downarrow 61\%$)} & 4903{\color{blue}($\uparrow$1.88$\times$)} \\
CF-ViT & 0.75 & 80.7{\color{blue}($+0.9$)} & 2.6{\color{blue}($\downarrow 43\%$)} & 3701{\color{blue}($\uparrow$1.32$\times$)} \\
CF-ViT & 1.0 & 80.8{\color{blue}($+1.0$)} & 4.0{\color{blue}($\downarrow 13\%$)} & 2760{\color{blue}($\uparrow$1.06$\times$)} \\
\hline
LV-ViT-S &- & 83.3 & 6.6 & 1681 \\
\hline
CF-ViT & 0.63 & 83.3{\color{blue}($+0.0$)} & 3.1{\color{blue}($\downarrow 53\%$)} & 3393{\color{blue}($\uparrow$2.01$\times$)}  \\
CF-ViT & 0.75 & 83.5{\color{blue}($+0.2$)} & 4.0{(\color{blue}$\downarrow 39\%$)} & 2827{\color{blue}($\uparrow$1.68$\times$)}  \\
CF-ViT & 1.0 & 83.6{\color{blue}($+0.3$)}&
6.1{\color{blue}($\downarrow 7\%$)} & 2022{\color{blue}($\uparrow$1.31$\times$)} \\
\hline
\end{tabular}
\end{table}

All training settings of our CF-ViT, such as image processing, learning rate, \emph{etc}, are to follow these of DeiT and LV-ViT. 
%
In the training phase, only conducting the fine-grained splitting at informative regions would affect the convergence. Therefore, we split the entire image into fine-grained patches in the first 200 epochs, and select informative coarse patches for fine-grained splitting in the remaining training process.
Our CF-ViT model is trained on a workstation with 4 A100 GPUs. Notably, both coarse stage and fine stage share the same network parameters. Due to different sizes of patches between coarse stage and fine stage, we downsample the coarse patches to the shape of fine-grained patches in order to facilitate sharing parameters in the patch embedding layer.

%
%

\subsection{Experimental Results}

\textbf{Model Efficiency}. 
To demonstrate our model efficiency, we conduct comparisons between our CF-ViT and its backbones. The measurement metrics include top-1 classification accuracy, model FLOPs and model throughput.
Following existing studies~\cite{wang2021not,liang2022evit}, the model throughput is measured as the number of processed images per second on a single  A100 GPU. 
We feed the model 50,000 images in the validation set of ImageNet with a batch size of 1,024, and record the total inference time. Then, the throughput is computed as $\frac{50,000}{\text{total inference time}}$.

\begin{table}[!t]
\centering
\caption{\label{token_slimming_compare_table} Comparisons between existing token slimming based ViT compression methods and our CF-ViT. }
\resizebox{\linewidth}{!}{
\begin{tabular}{ccc}
\hline\noalign{\smallskip}
Model & Top-1 Acc.(\%) & FLOPs(G) \\
\hline
\multicolumn{3}{c}{DeiT-S} \\
\hline
Baseline~\cite{touvron2021training} & 79.8 & 4.6 \\
DynamicViT~\cite{rao2021dynamicvit} & 79.3 & 2.9 \\
IA-RED$^2$~\cite{pan2021ia} & 79.1 & 3.2 \\
PS-ViT~\cite{tang2022patch} & 79.4 & 2.6 \\
EVIT~\cite{liang2022evit} & 79.5 & 3.0 \\
Evo-ViT~\cite{xu2022evovit} & 79.4 & 3.0 \\
\textbf{CF-ViT($\eta=0.5$)(Ours)} & \textbf{79.8} & \textbf{1.8} \\
\textbf{CF-ViT($\eta=0.75$)(Ours)} & \textbf{80.7} & \textbf{2.6} \\
\hline
\multicolumn{3}{c}{LV-ViT-S} \\
\hline
Baseline~\cite{jiang2021all} & 83.3 & 6.6 \\
DynamicViT~\cite{rao2021dynamicvit} & 83.0 & 4.6 \\
EVIT~\cite{liang2022evit} & 83.0 & 4.7 \\
SiT~\cite{zong2021self} & 83.2 & 4.0 \\
\textbf{CF-ViT($\eta=0.63$)(Ours)} & \textbf{83.3} & \textbf{3.1} \\
\textbf{CF-ViT($\eta=0.75$)(Ours)} & \textbf{83.5} & \textbf{4.0} \\
\hline
\end{tabular}
}
\end{table}

Table\,\ref{throughput_table} displays the comparison results
with varying values of threshold $\eta$ which balances accuracy and efficiency as discussed in Sec.\,\ref{training_strategies}. It can be observed that when maintaining the same accuracy with the backbone, CF-ViT significantly reduces model FLOPs of DeiT-S by 61\% and LV-ViT-S by 53\%. 
Consequently, our CF-ViT obtains a great power to process images, leading to 1.88$\times$ throughput improvements over DeiT-S and 2.01$\times$ over LV-ViT-S.
The supreme efficiency is attributed to our design of informative region identification which further splits only informative patches in the coarse stage.
Besides, with a larger $\eta$, our CF-ViT manifests not only FLOPs and throughput, but better top-1 accuracy. 
When $\eta = 1$ which indicates all the input are sent to fine inference stage, our CF-ViT significantly increases the performance of DeiT-S by 1.0\% and LV-ViT-S by 0.3\%. These results well demonstrate that our CF-ViT can well maintain a trade-off between model performance and model efficiency.


\textbf{Comparison with Compressed Models}. To demonstrate the efficacy of our coarse-to-fine patch splitting in reducing model complexity, we further compare our CF-ViT with recent studies on compressing ViT models, including token slimming compression and early-exiting compression.

\begin{figure}
\centering
\includegraphics[width=0.9\linewidth]{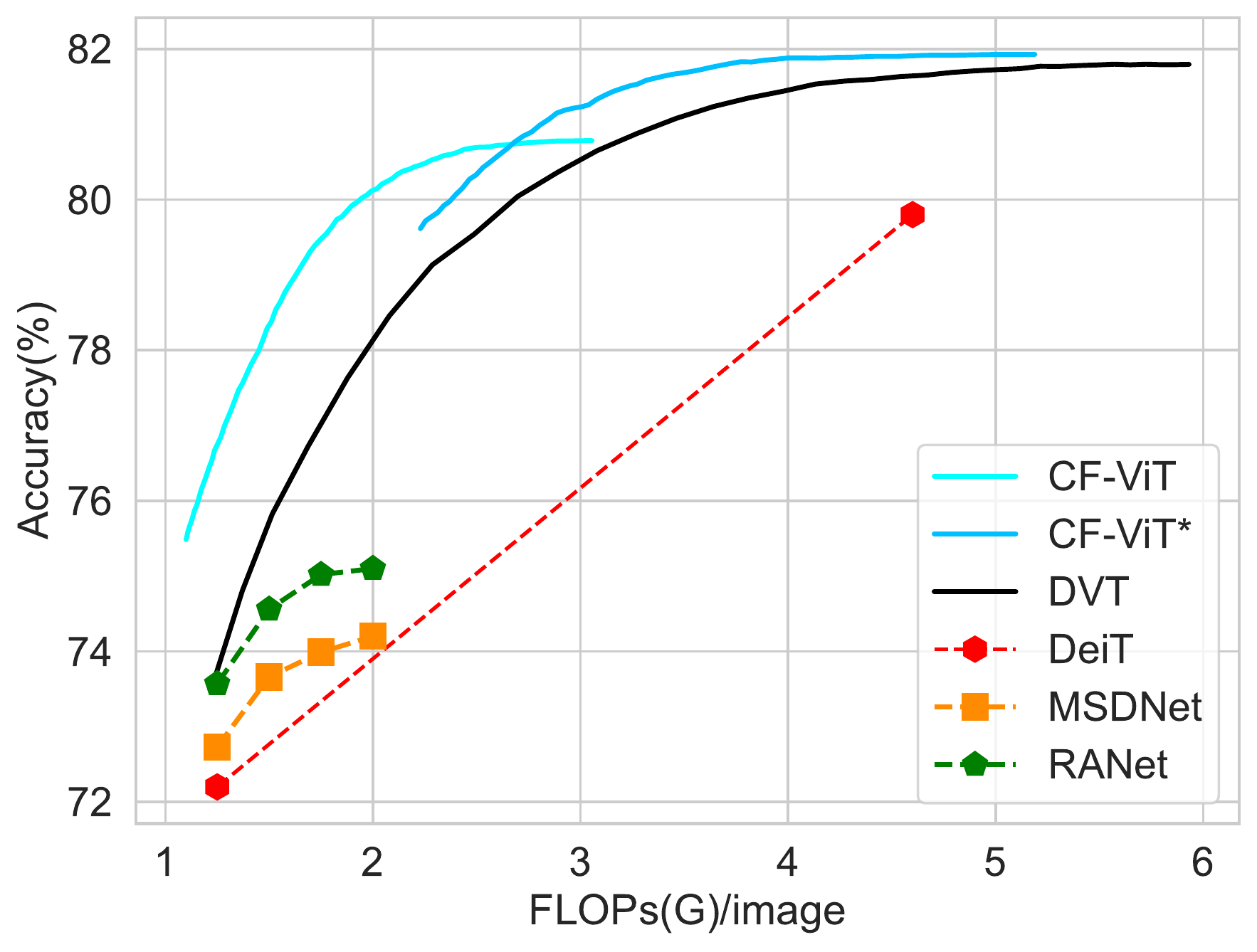}
\caption{\label{early_exit_compare_fig} Comparison between our CF-ViT and existing early-exiting methods. MSDNet~\cite{huang2017multi} and RANet~\cite{yang2020resolution} are CNN-based models. DVT~\cite{wang2021not} and CF-ViT are built upon DeiT.}
\end{figure}

(1) Token slimming compression reduces the complexity of ViT models by reducing the number of input tokens (patches), which is also the focus of this paper.
Table\,\ref{token_slimming_compare_table} shows the comparison with existing token slimming based ViT compression methods, including DynamicViT~\cite{rao2021dynamicvit}, IA-RED$^2$~\cite{pan2021ia}, PS-ViT~\cite{tang2022patch}, EViT~\cite{liang2022evit}, Evo-ViT~\cite{xu2022evovit} and SiT~\cite{zong2021self}. We report top-1 accuracy and FLOPs for performance evaluation. Results with DeiT-S and LV-ViT-S as backbones indicate that our CF-ViT outperforms previous methods \emph{w.r.t.} accuracy performance and FLOPs reduction. For example, CF-ViT significantly reduces the FLOPs of DeiT-S to 1.8G FLOPs without any compromise on accuracy performance, while the performance of recent advance, Evo-ViT, has only 79.4\% with much heavy FLOPs burden of 3.0G. Similar results can be observed when using LV-ViT-S as the backbone.

(2) Early-exiting compression stops the inference if the intermediate representation of an input satisfies a particular criterion, which is also considered in the coarse inference stage of our CF-ViT where the computational graph stops if the prediction confidence $p_j^c$ exceeds the threshold $\eta$.
In Fig.\,\ref{early_exit_compare_fig}, we further compare with the early-exiting methods, including CNN-based models such as MSDNet~\cite{huang2017multi} and RANet~\cite{yang2020resolution}, as well as transformer-based models such as DVT~\cite{wang2021not}. For fair comparison, 
both DVT and our CF-ViT are constructed upon DeiT-S as the backbone. From Fig.\,\ref{early_exit_compare_fig}, two phenomena can be observed: (1) Transformer-based models usually show supreme performance over CNN-based methods under similar FLOPs consumption. (2) Our CF-ViT consistently results in best accuracy than DVT that cascades multiple ViTs with an increasing token number. In contrast, our CF-ViT implements fine-grained splitting only for the informative regions, leading to an overall reduction in tokens. Thus, with similar accuracy, CF-ViT manifests smaller FLOPs consumption.

\begin{figure}[!t]
\centering
\includegraphics[width=0.9\linewidth]{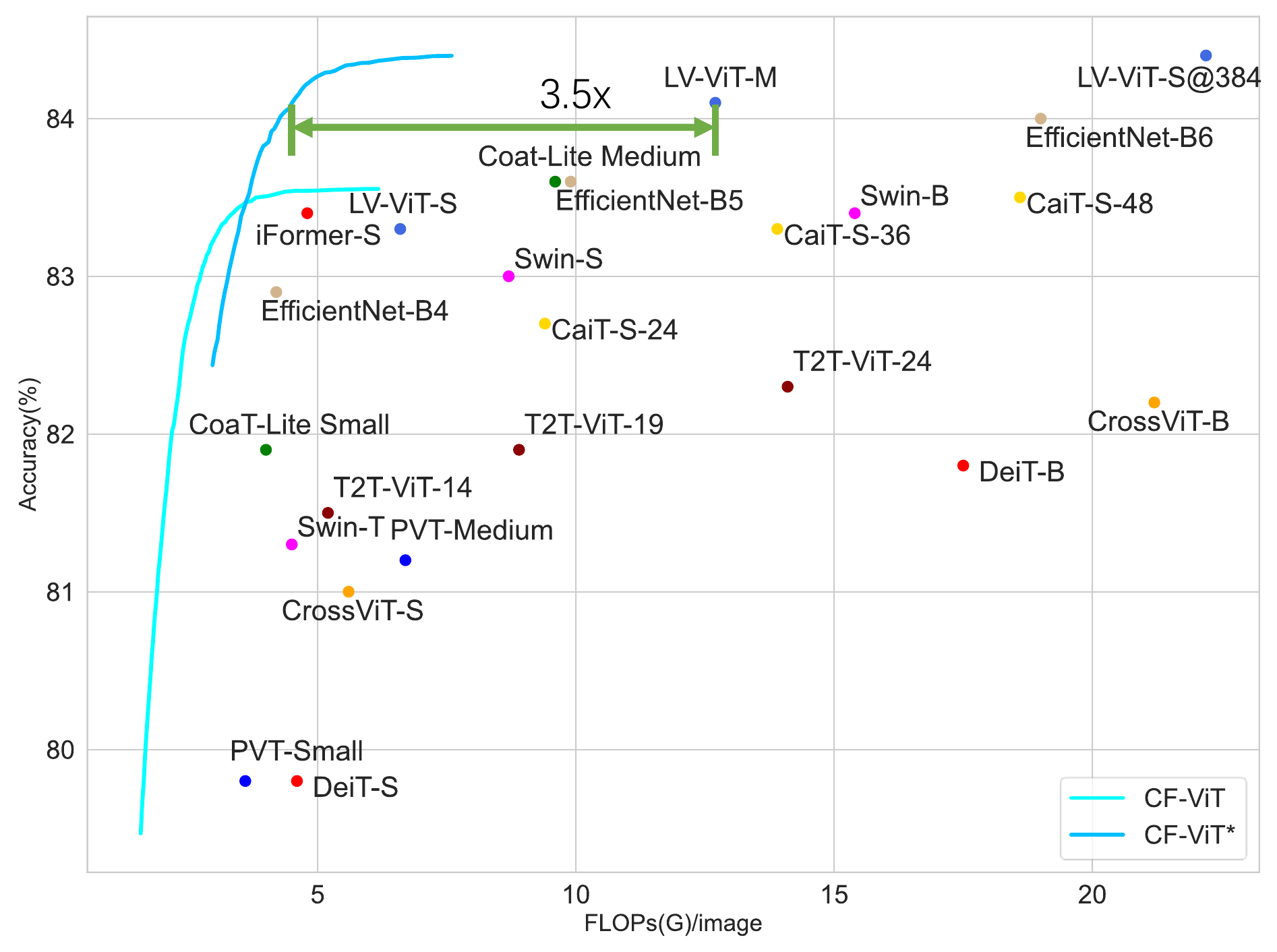}
\caption{\label{sota_compare_fig} Comparison with popular ViT models. Our CF-ViT is built upon LV-ViT-S.}
\end{figure}

\textbf{Comparison with SOTAs}.
Fig.\,\ref{sota_compare_fig} compares the accuracy and FLOPs trade-off of popular ViT models as well as our CF-ViT built upon LV-ViT-S\cite{jiang2021all}. The compared methods include DeiT~\cite{touvron2021training}, PVT~\cite{wang2021pyramid}, CoaT~\cite{xu2021co}, CrossViT~\cite{chen2021crossvit}, Swin~\cite{liu2021swin}, T2T-ViT~\cite{yuan2021tokens}, CaiT~\cite{touvron2021going}, iFormers~\cite{si2022inception}, and EfficientNet~\cite{tan2019efficientnet}. It can be observed that CF-ViT is significantly competitive in computation-accuracy trade-off among these baselines. For example, CF-ViT$^*$ achieves 84.1\% accuracy with $3.5\times$ less FLOPs compared with the vanilla LV-ViT-M.

\begin{table}[t]
\centering
\caption{\label{selection_ablation_table} Performance comparison between our informative region identification and its variants. GCA means global class attention. }
\begin{tabular}{ccc}
\hline\noalign{\smallskip}
\multirow{2}{*}{Ablation} & \multicolumn{2}{c}{Top-1 Acc.(\%)} \\
 & coarse & fine \\
\hline
negative GCA & 74.9 & 77.6 \\
random & 75.3 & 79.6 \\
last class attention & 75.3 & 80.3 \\
\hline
\textbf{Ours} & \textbf{75.5} & \textbf{80.8} \\
\hline
\end{tabular}
\end{table}

\begin{table}[t]
\centering
\caption{\label{reuse_ablation_table} Performance comparison between our feature reuse and its variants.}
\begin{tabular}{ccc}
\hline\noalign{\smallskip}
\multirow{2}{*}{Ablation} & \multicolumn{2}{c}{Top-1 Acc.(\%)} \\
 & coarse & fine  \\
\hline
w/o reuse & 75.2 & 80.0  \\
Ours + [class] token & 75.4 & 80.2 \\
Ours + uninformative tokens & 75.4 & 80.6  \\
MLP$\to$Linear & 75.3 & 80.6 \\
\hline
\textbf{Ours} & \textbf{75.5} & \textbf{80.8} \\
\hline
\end{tabular}
\end{table}

\subsection{Ablation Study}
We analyze the efficacy of each design in our CF-ViT, including informative region identification, feature reuse and early-exiting. To show the effectiveness of our designs, we also compare our informative region identification and feature reuse with other alternatives. To show their necessity, we remove each design individually and display the performance. All ablation studies take DeiT-S as backbone.

\textbf{Informative region identification}.
Three variants are developed to replace our informative region identification: (1) Negative global class attention, which selects the regions with smaller global class attention. (2) Random, which randomly picks up regions for fine-grained. (3) Last class attention, which leverages the class attention in the last encoder. We deactivate the early termination, and compare the performance in Table\,\ref{selection_ablation_table}. It is intuitive that the negative global class attention results in the poorest performance since the most informative are removed in this setting. This result is in coincidence with the observation in Fig.\,\ref{class_attention_fig}(a). Our informative region identification considers the global class attention, leading to performance increase compared to this only considering the last class attention, well demonstrating the correctness of our motive to combine class attention across different layers.


%
\textbf{Feature Reuse}.
Our feature reuse leverages a MLP to process the output image token in the coarse inference stage and shortcuts them to the fine inference stage. Note that we do not reuse the [class] token and uninformative tokens. 
Table\,\ref{reuse_ablation_table} compares our feature reuse with three variants including: (1) Integrating [class] token to our feature reuse. (2) Integrating uninformative tokens to our feature reuse. (3) Replacing the MLP with one single linear layer. 
From Table\,\ref{reuse_ablation_table}, we can see that considering [class] token or uninformative tokens has a negative impact on the performance. Besides, replacing the MLP layer with a linear layer drops down the performance from 80.8\% to 80.6\%. These results well demonstrate the effectiveness of our design of feature reuse.



\begin{figure}[!t]
\centering
\includegraphics[width=0.9\linewidth]{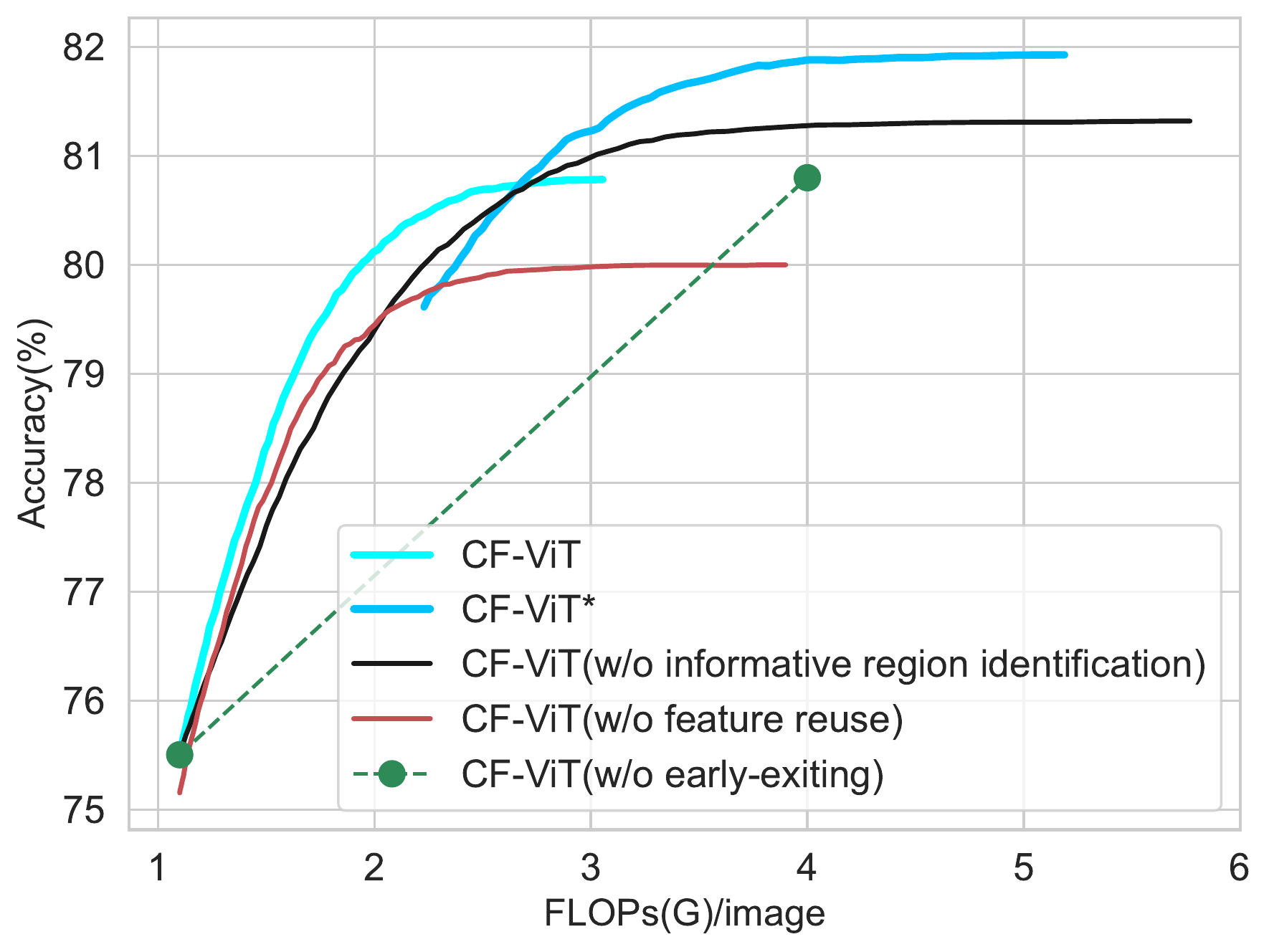}
\caption{\label{method_ablation} Performance analysis of removing each of the three designs.}   
\end{figure}

\begin{table}[!t]
\caption{Accuracy and FLOPs with different values of $\alpha$.}
\centering
\setlength{\tabcolsep}{1mm}
\begin{tabular}{ccccccc}
\hline
 & 0.4 & \textbf{0.5(default)} & 0.6 & 0.7 & 0.8 & 0.9  \\ \hline
Top1-Acc(\%) & 80.4 & 80.8 & 80.9 & 81.1 & 81.3 & 81.4 \\ \hline
FLOPs(G) & 3.7 & 4.0 & 4.4 & 4.7 & 5.1 & 5.5   \\ \hline
\end{tabular}
\label{different_alpha_table}
\end{table}

\begin{table}[!t]
\caption{Accuracy with different values of $\beta$.}
\centering
\setlength{\tabcolsep}{1mm}
\begin{tabular}{ccccccc}
\hline
  & 0 & 0.5 & 0.9 & \textbf{0.99(default)} & 0.999   \\ \hline
Top1-Acc(\%) & 80.3 & 80.5 & 80.7 & 80.8 & 80.8  \\ \hline
\end{tabular}
\label{different_beta_table}
\end{table}

\begin{table}[t]
\centering
\caption{\label{loss_ablation_table} Performance comparison between different loss function. }
\begin{tabular}{ccc}
\hline\noalign{\smallskip}
\multirow{2}{*}{Ablation} & \multicolumn{2}{c}{Top-1 Acc.(\%)} \\
 & coarse & fine \\
\hline
CE + CE & \textbf{75.7} & 80.3 \\
\hline
\textbf{CE + KL(ours)} & 75.5 & \textbf{80.8} \\
\hline
\end{tabular}
\end{table}

\textbf{Necessity of each design}.
Fig.\,\ref{method_ablation} plots the performance of our CF-ViT by individually removing each design. Generally, we can observe that the removal of each component incurs severe performance drops. Thus, all three designs are vital to the final performance of our CF-ViT.
It is worth noting that, the one without informative region identification split all the coarse-grained patches, leading to a drastic increase of tokens. However, our informative identification chooses to split only informative regions, which greatly reduces tokens and thus brings less computational cost.

\textbf{Influence of $\alpha$}.
Tab.\,\ref{different_alpha_table} provides accuracy of fine inference stage and FLOPs 
with different values of $\alpha$ \big(see Eq.\,\ref{input_num_equal}\big).  
We can see that from Tab.\,\ref{different_alpha_table} that,
a larger $\alpha$ leads to better accuracy but more FLOPs consumption. In this paper, we set $\alpha$ as 0.5 for a accuracy-FLOPs trade-off.

\textbf{Influence of $\beta$}.
Tab. \ref{different_beta_table} provides accuracy of fine inference stage with different values of $\beta$ \big(see Eq.\,\ref{global_class_attention_equal}\big).  
It is intuitive that $\beta$ indicates the weight of attention from the shallow encoder. 
The $\beta=0$ indicates that only the class attention from the last encoder is used. We set $\beta=0.99$ as default for its optimal performance.

\textbf{Influence of loss function}.
As shown in Eq.\,(\ref{loss_equation}), we use $CE(\cdot,\cdot)$ to make the output of fine stage fit the truth label, and use $KL(\cdot,\cdot)$ to make the output of coarse stage fit the output of fine stage. We also try to make the output of coarse stage and the output of fine stage both fit the truth label:
\begin{equation}\label{loss_equation_2}
    \hat{loss} = CE(\mathbf{p}_f, \mathbf{y}) + CE(\mathbf{p}_c, \mathbf{y}),
\end{equation}
As show in Tab.\,\ref{loss_ablation_table}, compare with origin CE+KL \big(Eq.\,(\ref{loss_equation})\big), CE+CE \big(Eq.\,(\ref{loss_equation_2})\big) implement cause slightly benefit (+0.2\%) in coarse inference stage but more degradation (-0.5\%) in fine inference stage. We choose Eq.\,(\ref{loss_equation}) as the loss function because of the significant benefits in fine inference stage.

\begin{figure}[!t]
\centering
\includegraphics[width=0.9\linewidth]{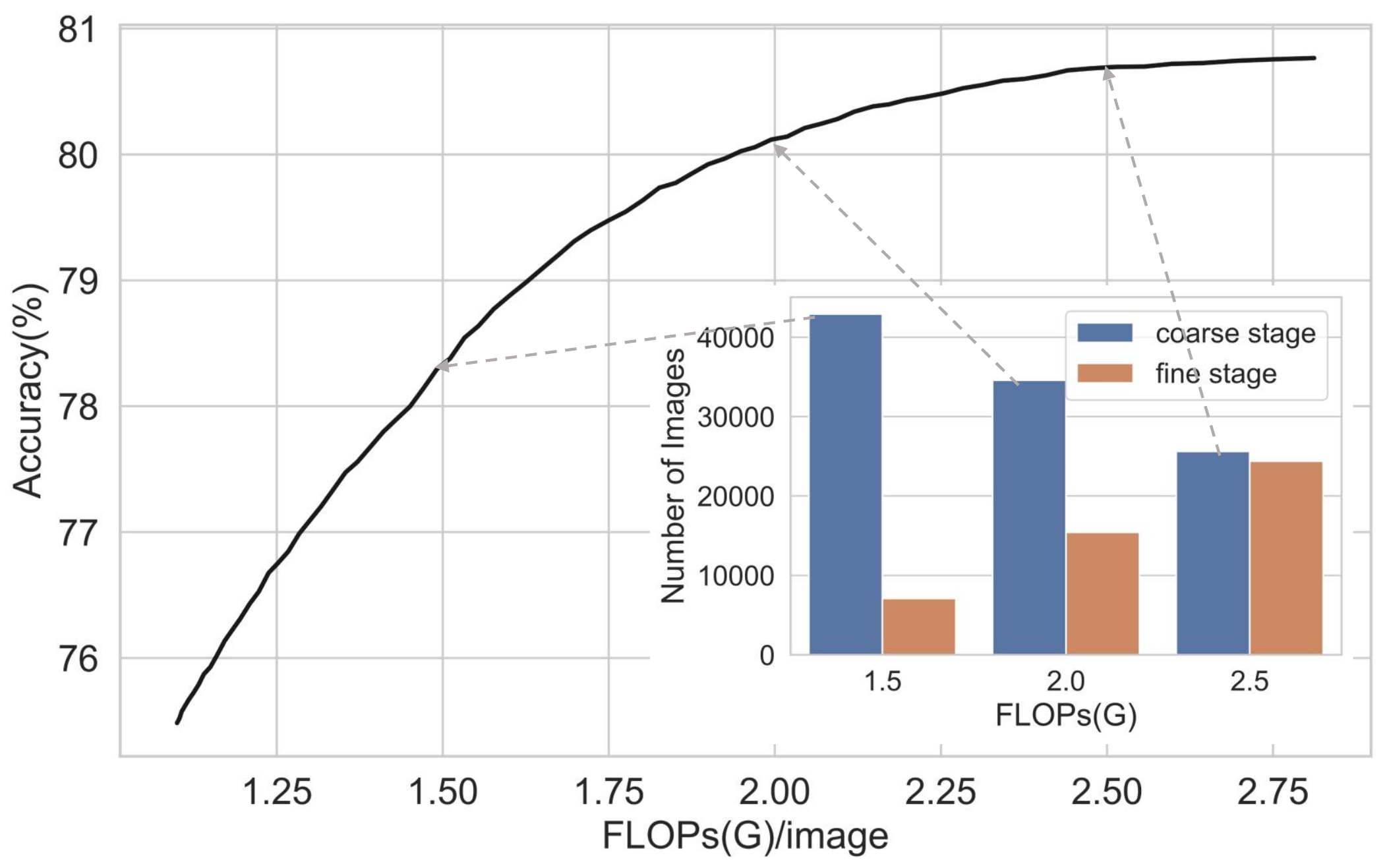}
\caption{\label{exit_num_fig} No. of images correctly classified at coarse and fine stages.}
\end{figure}

\begin{figure*}[!t]
\centering
\includegraphics[width=0.85\textwidth]{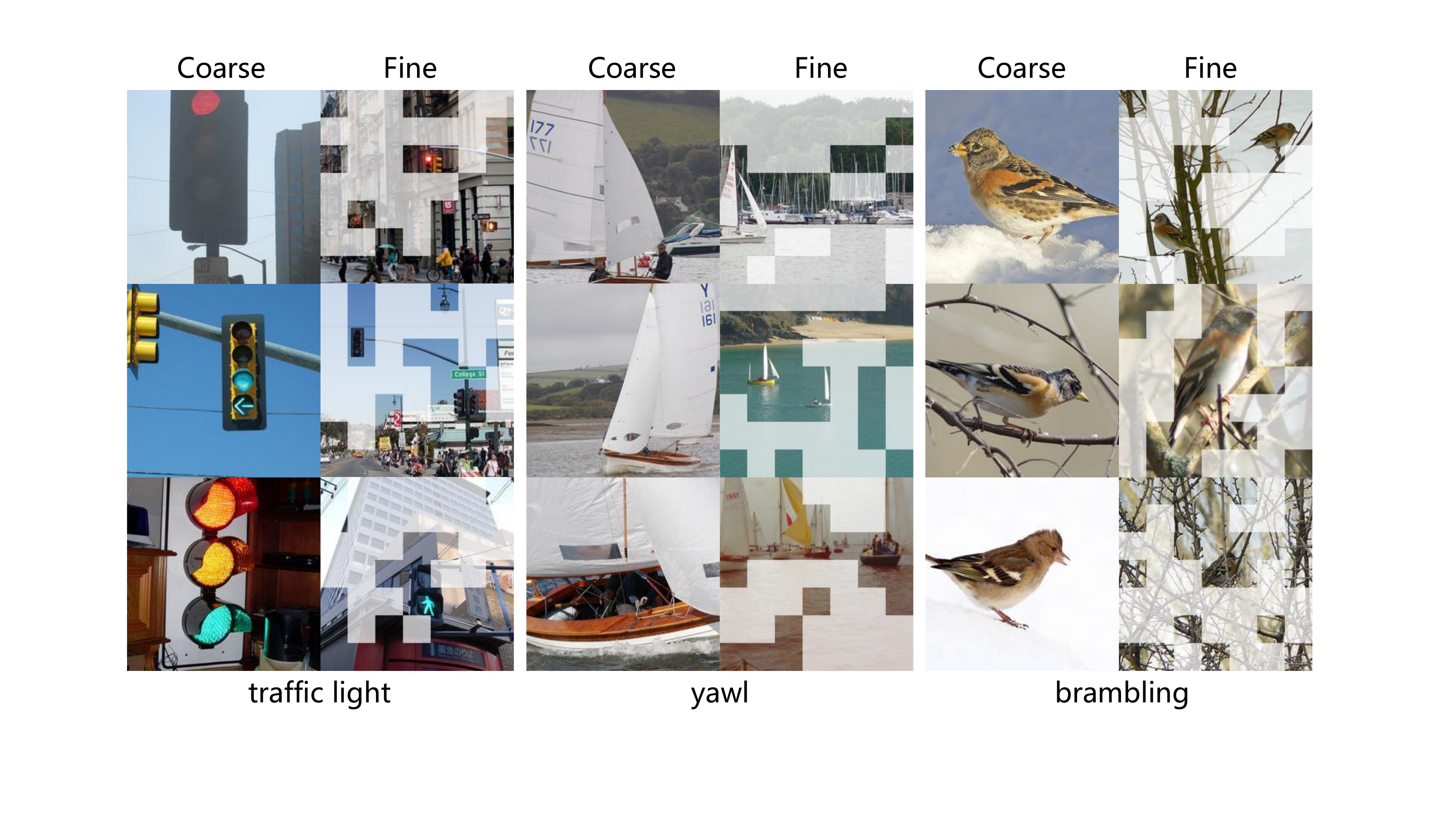}
\caption{\label{visulization_fig}Illustration of images correctly classified at coarse stage and fine stage. For fine stage, we visualize the regions selected by our informative region identification (grey boxes) indicate the uninformative patches. }
\end{figure*}

\subsection{Visualization}
In Fig.\,\ref{visulization_fig}, we illustrate some images that are correctly recognized at coarse inference by CF-ViT(DeiT-S), as well as some recognized at fine inference stage. For a better illustration, we only visualize informative regions if images are recognized at fine inference stage.
We can observe an overall trend that images well classified at coarse stage are mostly filled with ``easy'' regions. Consequently, a coarse-grained splitting can well tell the categories of these images. On the contrary, these containing complex scenes and obscure objects require to be further split for a correct recognition, which is realized at fine stage. Besides, the selected regions by our informative region identification mostly locate the target objects. In Fig.\,\ref{exit_num_fig}, we also show some statistics \emph{w.r.t.} the number of images correctly classified at coarse stage and fine stage. By adjusting the threshold $\eta$, CF-ViT models can be obtained with different computational budgets. With a larger $\eta$, more images will be further split and fed to fine inference stage for recognition. Therefore, the increasing $\eta$ brings about more computational cost and the number of images correctly classified at fine stage also increases.

\section{Conclusion}
This paper focuses on reducing the redundant input tokens for accelerating vision transformers.  Specially, we proposed a coarse-to-fine vision transformer (CF-ViT), the inference of which is two-fold including a coarse inference stage and a fine inference stage. The former splits the input image into a small-length token sequence to recognize these images filled with ``easy'' regions in a computationally economical manner while the latter further splits the informative patches for a better recognition if the coarse inference does not well classify the input. Extensive experiments indicate that our CF-ViT can achieve a better trade-off between performance and efficiency. 
Furthermore, transferring the coarse-to-fine inference paradigm to 
dense prediction tasks such as object detection and semantic segmentation will be included in our future work.


\section*{Acknowledgement}
\begin{sloppypar}
This work is supported by the National Science Fund for Distinguished Young Scholars (No. 62025603), the National Natural Science Foundation of China (No. U1705262, No. 62176222, No. 62176223, No. 62176226, No. 62072386, No. 62072387, No. 62072389, No. 62002305, No. 61772443, No. 61802324 and No. 61702136),
Guangdong Basic and Applied Basic Research Foundation (No.2019B1515120049), the Natural Science Foundation of Fujian Province of China (No. 2021J01002), and the Fundamental Research Funds for the central universities (No. 20720200077, No. 20720200090 and No. 20720200091).
\end{sloppypar}

\appendix

\bibliography{aaai23}

\begin{thebibliography}{48}
\providecommand{\natexlab}[1]{#1}

\bibitem[{Chen, Fan, and Panda(2021)}]{chen2021crossvit}
Chen, C.-F.~R.; Fan, Q.; and Panda, R. 2021.
\newblock Crossvit: Cross-attention multi-scale vision transformer for image
  classification.
\newblock In \emph{Proceedings of the IEEE/CVF International Conference on
  Computer Vision (ICCV)}, 357--366.

\bibitem[{Chu et~al.(2021{\natexlab{a}})Chu, Tian, Wang, Zhang, Ren, Wei, Xia,
  and Shen}]{chu2021twins}
Chu, X.; Tian, Z.; Wang, Y.; Zhang, B.; Ren, H.; Wei, X.; Xia, H.; and Shen, C.
  2021{\natexlab{a}}.
\newblock Twins: Revisiting the design of spatial attention in vision
  transformers.
\newblock In \emph{Advances in Neural Information Processing Systems
  (NeurIPS)}.

\bibitem[{Chu et~al.(2021{\natexlab{b}})Chu, Tian, Zhang, Wang, Wei, Xia, and
  Shen}]{chu2021conditional}
Chu, X.; Tian, Z.; Zhang, B.; Wang, X.; Wei, X.; Xia, H.; and Shen, C.
  2021{\natexlab{b}}.
\newblock Conditional positional encodings for vision transformers.
\newblock arXiv:2102.10882.

\bibitem[{Deng et~al.(2009)Deng, Dong, Socher, Li, Li, and
  Fei-Fei}]{deng2009imagenet}
Deng, J.; Dong, W.; Socher, R.; Li, L.-J.; Li, K.; and Fei-Fei, L. 2009.
\newblock Imagenet: A large-scale hierarchical image database.
\newblock In \emph{Proceedings of the IEEE/CVF International Conference on
  Computer Vision (ICCV)}, 248--255.

\bibitem[{Dosovitskiy et~al.(2020)Dosovitskiy, Beyer, Kolesnikov, Weissenborn,
  Zhai, Unterthiner, Dehghani, Minderer, Heigold, Gelly
  et~al.}]{dosovitskiy2020image}
Dosovitskiy, A.; Beyer, L.; Kolesnikov, A.; Weissenborn, D.; Zhai, X.;
  Unterthiner, T.; Dehghani, M.; Minderer, M.; Heigold, G.; Gelly, S.; et~al.
  2020.
\newblock An Image is Worth 16x16 Words: Transformers for Image Recognition at
  Scale.
\newblock In \emph{International Conference on Learning Representations
  (ICLR)}.

\bibitem[{Fang et~al.(2021)Fang, Xie, Wang, Zhang, Liu, and Tian}]{fang2021msg}
Fang, J.; Xie, L.; Wang, X.; Zhang, X.; Liu, W.; and Tian, Q. 2021.
\newblock Msg-transformer: Exchanging local spatial information by manipulating
  messenger tokens.
\newblock arXiv:2105.15168.

\bibitem[{Han et~al.(2022{\natexlab{a}})Han, Wang, Chen, Chen, Guo, Liu, Tang,
  Xiao, Xu, Xu et~al.}]{han2022survey}
Han, K.; Wang, Y.; Chen, H.; Chen, X.; Guo, J.; Liu, Z.; Tang, Y.; Xiao, A.;
  Xu, C.; Xu, Y.; et~al. 2022{\natexlab{a}}.
\newblock A survey on vision transformer.
\newblock \emph{IEEE Transactions on Pattern Analysis and Machine Intelligence
  (TPAMI)}.

\bibitem[{Han et~al.(2021{\natexlab{a}})Han, Xiao, Wu, Guo, Xu, and
  Wang}]{han2021transformer}
Han, K.; Xiao, A.; Wu, E.; Guo, J.; Xu, C.; and Wang, Y. 2021{\natexlab{a}}.
\newblock Transformer in transformer.
\newblock In \emph{Advances in Neural Information Processing Systems
  (NeurIPS)}.

\bibitem[{Han et~al.(2021{\natexlab{b}})Han, Huang, Song, Yang, Wang, and
  Wang}]{han2021dynamic}
Han, Y.; Huang, G.; Song, S.; Yang, L.; Wang, H.; and Wang, Y.
  2021{\natexlab{b}}.
\newblock Dynamic neural networks: A survey.
\newblock \emph{IEEE Transactions on Pattern Analysis and Machine Intelligence
  (TPAMI)}.

\bibitem[{Han et~al.(2021{\natexlab{c}})Han, Huang, Song, Yang, Zhang, and
  Jiang}]{han2021spatially}
Han, Y.; Huang, G.; Song, S.; Yang, L.; Zhang, Y.; and Jiang, H.
  2021{\natexlab{c}}.
\newblock Spatially adaptive feature refinement for efficient inference.
\newblock \emph{IEEE Transactions on Image Processing}, 9345--9358.

\bibitem[{Han et~al.(2022{\natexlab{b}})Han, Yuan, Pu, Xue, Song, Sun, and
  Huang}]{han2022latency}
Han, Y.; Yuan, Z.; Pu, Y.; Xue, C.; Song, S.; Sun, G.; and Huang, G.
  2022{\natexlab{b}}.
\newblock Latency-aware Spatial-wise Dynamic Networks.
\newblock In \emph{Advances in Neural Information Processing Systems
  (NeurIPS)}.

\bibitem[{Heo et~al.(2021)Heo, Yun, Han, Chun, Choe, and
  Oh}]{heo2021rethinking}
Heo, B.; Yun, S.; Han, D.; Chun, S.; Choe, J.; and Oh, S.~J. 2021.
\newblock Rethinking spatial dimensions of vision transformers.
\newblock In \emph{Proceedings of the IEEE/CVF International Conference on
  Computer Vision (ICCV)}, 11936--11945.

\bibitem[{Huang et~al.(2018)Huang, Chen, Li, Wu, van~der Maaten, and
  Weinberger}]{huang2017multi}
Huang, G.; Chen, D.; Li, T.; Wu, F.; van~der Maaten, L.; and Weinberger, K.~Q.
  2018.
\newblock Multi-scale dense networks for resource efficient image
  classification.
\newblock In \emph{International Conference on Learning Representations
  (ICLR)}.

\bibitem[{Huang et~al.(2021)Huang, Ben, Luo, Cheng, Yu, and
  Fu}]{huang2021shuffle}
Huang, Z.; Ben, Y.; Luo, G.; Cheng, P.; Yu, G.; and Fu, B. 2021.
\newblock Shuffle transformer: Rethinking spatial shuffle for vision
  transformer.
\newblock arXiv:2106.03650.

\bibitem[{Jiang et~al.(2021)Jiang, Hou, Yuan, Zhou, Shi, Jin, Wang, and
  Feng}]{jiang2021all}
Jiang, Z.-H.; Hou, Q.; Yuan, L.; Zhou, D.; Shi, Y.; Jin, X.; Wang, A.; and
  Feng, J. 2021.
\newblock All tokens matter: Token labeling for training better vision
  transformers.
\newblock In \emph{Advances in Neural Information Processing Systems
  (NeurIPS)}.

\bibitem[{Li et~al.(2022)Li, Wang, Liu, Tan, Lin, Wu, Chen, Zheng, and
  Li}]{li2022efficient}
Li, S.; Wang, Z.; Liu, Z.; Tan, C.; Lin, H.; Wu, D.; Chen, Z.; Zheng, J.; and
  Li, S.~Z. 2022.
\newblock Efficient Multi-order Gated Aggregation Network.
\newblock arXiv:2211.03295.

\bibitem[{Li et~al.(2021)Li, Zhang, Cao, Timofte, and
  Van~Gool}]{li2021localvit}
Li, Y.; Zhang, K.; Cao, J.; Timofte, R.; and Van~Gool, L. 2021.
\newblock Localvit: Bringing locality to vision transformers.
\newblock arXiv:2104.05707.

\bibitem[{Liang et~al.(2022)Liang, GE, Tong, Song, Wang, and
  Xie}]{liang2022evit}
Liang, Y.; GE, C.; Tong, Z.; Song, Y.; Wang, J.; and Xie, P. 2022.
\newblock {E}viT: Expediting Vision Transformers via Token Reorganizations.
\newblock In \emph{International Conference on Learning Representations
  (ICLR)}.

\bibitem[{Lin et~al.(2022)Lin, Chen, Zhang, Li, Shen, Shen, and
  Ji}]{lin2022super}
Lin, M.; Chen, M.; Zhang, Y.; Li, K.; Shen, Y.; Shen, C.; and Ji, R. 2022.
\newblock Super Vision Transformer.
\newblock arXiv:2205.11397.

\bibitem[{Liu et~al.(2021)Liu, Lin, Cao, Hu, Wei, Zhang, Lin, and
  Guo}]{liu2021swin}
Liu, Z.; Lin, Y.; Cao, Y.; Hu, H.; Wei, Y.; Zhang, Z.; Lin, S.; and Guo, B.
  2021.
\newblock Swin transformer: Hierarchical vision transformer using shifted
  windows.
\newblock In \emph{Proceedings of the IEEE/CVF International Conference on
  Computer Vision (ICCV)}, 10012--10022.

\bibitem[{Pan et~al.(2021{\natexlab{a}})Pan, Panda, Jiang, Wang, Feris, and
  Oliva}]{pan2021ia}
Pan, B.; Panda, R.; Jiang, Y.; Wang, Z.; Feris, R.; and Oliva, A.
  2021{\natexlab{a}}.
\newblock IA-RED$^{2}$: Interpretability-Aware Redundancy Reduction for Vision
  Transformers.
\newblock In \emph{Advances in Neural Information Processing Systems
  (NeurIPS)}.

\bibitem[{Pan et~al.(2021{\natexlab{b}})Pan, Zhuang, Liu, He, and
  Cai}]{pan2021scalable}
Pan, Z.; Zhuang, B.; Liu, J.; He, H.; and Cai, J. 2021{\natexlab{b}}.
\newblock Scalable vision transformers with hierarchical pooling.
\newblock In \emph{Proceedings of the IEEE/CVF International Conference on
  Computer Vision (ICCV)}, 377--386.

\bibitem[{Rao et~al.(2021)Rao, Zhao, Liu, Lu, Zhou, and
  Hsieh}]{rao2021dynamicvit}
Rao, Y.; Zhao, W.; Liu, B.; Lu, J.; Zhou, J.; and Hsieh, C.-J. 2021.
\newblock Dynamicvit: Efficient vision transformers with dynamic token
  sparsification.
\newblock In \emph{Advances in Neural Information Processing Systems
  (NeurIPS)}.

\bibitem[{Si et~al.(2022)Si, Yu, Zhou, Zhou, Wang, and Yan}]{si2022inception}
Si, C.; Yu, W.; Zhou, P.; Zhou, Y.; Wang, X.; and Yan, S. 2022.
\newblock Inception Transformer.
\newblock In \emph{Advances in Neural Information Processing Systems
  (NeurIPS)}.

\bibitem[{Song et~al.(2021)Song, Zhang, Liu, Li, He, Sun, Sun, and
  Zheng}]{song2021dynamic}
Song, L.; Zhang, S.; Liu, S.; Li, Z.; He, X.; Sun, H.; Sun, J.; and Zheng, N.
  2021.
\newblock Dynamic grained encoder for vision transformers.
\newblock In \emph{Advances in Neural Information Processing Systems
  (NeurIPS)}.

\bibitem[{Sun et~al.(2017)Sun, Shrivastava, Singh, and
  Gupta}]{sun2017revisiting}
Sun, C.; Shrivastava, A.; Singh, S.; and Gupta, A. 2017.
\newblock Revisiting unreasonable effectiveness of data in deep learning era.
\newblock In \emph{Proceedings of the IEEE International Conference on Computer
  Vision (ICCV)}, 843--852.

\bibitem[{Tan and Le(2019)}]{tan2019efficientnet}
Tan, M.; and Le, Q. 2019.
\newblock Efficientnet: Rethinking model scaling for convolutional neural
  networks.
\newblock In \emph{International Conference on Machine Learning (ICML)},
  6105--6114.

\bibitem[{Tang et~al.(2022{\natexlab{a}})Tang, Zhang, Zhu, and
  Tan}]{tang2022quadtree}
Tang, S.; Zhang, J.; Zhu, S.; and Tan, P. 2022{\natexlab{a}}.
\newblock Quadtree Attention for Vision Transformers.
\newblock In \emph{International Conference on Learning Representations
  (ICLR)}.

\bibitem[{Tang et~al.(2022{\natexlab{b}})Tang, Han, Wang, Xu, Guo, Xu, and
  Tao}]{tang2022patch}
Tang, Y.; Han, K.; Wang, Y.; Xu, C.; Guo, J.; Xu, C.; and Tao, D.
  2022{\natexlab{b}}.
\newblock Patch slimming for efficient vision transformers.
\newblock In \emph{Proceedings of the IEEE/CVF Conference on Computer Vision
  and Pattern Recognition (CVPR)}, 12165--12174.

\bibitem[{Touvron et~al.(2021{\natexlab{a}})Touvron, Cord, Douze, Massa,
  Sablayrolles, and J{\'e}gou}]{touvron2021training}
Touvron, H.; Cord, M.; Douze, M.; Massa, F.; Sablayrolles, A.; and J{\'e}gou,
  H. 2021{\natexlab{a}}.
\newblock Training data-efficient image transformers \& distillation through
  attention.
\newblock In \emph{International Conference on Machine Learning (ICML)},
  10347--10357.

\bibitem[{Touvron et~al.(2021{\natexlab{b}})Touvron, Cord, Sablayrolles,
  Synnaeve, and J{\'e}gou}]{touvron2021going}
Touvron, H.; Cord, M.; Sablayrolles, A.; Synnaeve, G.; and J{\'e}gou, H.
  2021{\natexlab{b}}.
\newblock Going deeper with image transformers.
\newblock In \emph{Proceedings of the IEEE/CVF International Conference on
  Computer Vision (ICCV)}, 32--42.

\bibitem[{Vaswani et~al.(2017)Vaswani, Shazeer, Parmar, Uszkoreit, Jones,
  Gomez, Kaiser, and Polosukhin}]{vaswani2017attention}
Vaswani, A.; Shazeer, N.; Parmar, N.; Uszkoreit, J.; Jones, L.; Gomez, A.~N.;
  Kaiser, {\L}.; and Polosukhin, I. 2017.
\newblock Attention is all you need.
\newblock In \emph{Advances in Neural Information Processing Systems
  (NeurIPS)}.

\bibitem[{Wang, Stuijk, and De~Haan(2014)}]{wang2014exploiting}
Wang, W.; Stuijk, S.; and De~Haan, G. 2014.
\newblock Exploiting spatial redundancy of image sensor for motion robust rPPG.
\newblock \emph{IEEE Transactions on Biomedical Engineering (TBE)}, 415--425.

\bibitem[{Wang et~al.(2021{\natexlab{a}})Wang, Xie, Li, Fan, Song, Liang, Lu,
  Luo, and Shao}]{wang2021pyramid}
Wang, W.; Xie, E.; Li, X.; Fan, D.-P.; Song, K.; Liang, D.; Lu, T.; Luo, P.;
  and Shao, L. 2021{\natexlab{a}}.
\newblock Pyramid vision transformer: A versatile backbone for dense prediction
  without convolutions.
\newblock In \emph{Proceedings of the IEEE/CVF International Conference on
  Computer Vision (ICCV)}, 568--578.

\bibitem[{Wang et~al.(2021{\natexlab{b}})Wang, Chen, Jiang, Song, Han, and
  Huang}]{wang2021adaptive}
Wang, Y.; Chen, Z.; Jiang, H.; Song, S.; Han, Y.; and Huang, G.
  2021{\natexlab{b}}.
\newblock Adaptive focus for efficient video recognition.
\newblock In \emph{Proceedings of the IEEE/CVF International Conference on
  Computer Vision (ICCV)}, 16249--16258.

\bibitem[{Wang et~al.(2021{\natexlab{c}})Wang, Huang, Song, Huang, and
  Huang}]{wang2021not}
Wang, Y.; Huang, R.; Song, S.; Huang, Z.; and Huang, G. 2021{\natexlab{c}}.
\newblock Not all images are worth 16x16 words: Dynamic transformers for
  efficient image recognition.
\newblock In \emph{Advances in Neural Information Processing Systems
  (NeurIPS)}.

\bibitem[{Wang et~al.(2020)Wang, Lv, Huang, Song, Yang, and
  Huang}]{wang2020glance}
Wang, Y.; Lv, K.; Huang, R.; Song, S.; Yang, L.; and Huang, G. 2020.
\newblock Glance and focus: a dynamic approach to reducing spatial redundancy
  in image classification.
\newblock In \emph{Advances in Neural Information Processing Systems
  (NeurIPS)}, 2432--2444.

\bibitem[{Wang et~al.(2022{\natexlab{a}})Wang, Yue, Lin, Jiang, Lai, Kulikov,
  Orlov, Shi, and Huang}]{wang2022adafocus}
Wang, Y.; Yue, Y.; Lin, Y.; Jiang, H.; Lai, Z.; Kulikov, V.; Orlov, N.; Shi,
  H.; and Huang, G. 2022{\natexlab{a}}.
\newblock Adafocus v2: End-to-end training of spatial dynamic networks for
  video recognition.
\newblock In \emph{IEEE/CVF Conference on Computer Vision and Pattern
  Recognition (CVPR)}, 20030--20040.

\bibitem[{Wang et~al.(2022{\natexlab{b}})Wang, Yue, Xu, Hassani, Kulikov,
  Orlov, Song, Shi, and Huang}]{wang2022adafocusv3}
Wang, Y.; Yue, Y.; Xu, X.; Hassani, A.; Kulikov, V.; Orlov, N.; Song, S.; Shi,
  H.; and Huang, G. 2022{\natexlab{b}}.
\newblock AdaFocusV3: On Unified Spatial-Temporal Dynamic Video Recognition.
\newblock In \emph{European Conference on Computer Vision (ECCV)}, 226--243.

\bibitem[{Xie et~al.(2021)Xie, Wang, Yu, Anandkumar, Alvarez, and
  Luo}]{xie2021segformer}
Xie, E.; Wang, W.; Yu, Z.; Anandkumar, A.; Alvarez, J.~M.; and Luo, P. 2021.
\newblock SegFormer: Simple and efficient design for semantic segmentation with
  transformers.
\newblock In \emph{Advances in Neural Information Processing Systems
  (NeurIPS)}.

\bibitem[{Xu et~al.(2021)Xu, Xu, Chang, and Tu}]{xu2021co}
Xu, W.; Xu, Y.; Chang, T.; and Tu, Z. 2021.
\newblock Co-scale conv-attentional image transformers.
\newblock In \emph{Proceedings of the IEEE/CVF International Conference on
  Computer Vision (ICCV)}, 9981--9990.

\bibitem[{Xu et~al.(2022)Xu, Zhang, Zhang, Sheng, Li, Dong, Zhang, Xu, and
  Sun}]{xu2022evovit}
Xu, Y.; Zhang, Z.; Zhang, M.; Sheng, K.; Li, K.; Dong, W.; Zhang, L.; Xu, C.;
  and Sun, X. 2022.
\newblock Evo-ViT: Slow-Fast Token Evolution for Dynamic Vision Transformer.
\newblock In \emph{Proceedings of the AAAI Conference on Artificial
  Intelligence (AAAI)}.

\bibitem[{Yang et~al.(2020)Yang, Han, Chen, Song, Dai, and
  Huang}]{yang2020resolution}
Yang, L.; Han, Y.; Chen, X.; Song, S.; Dai, J.; and Huang, G. 2020.
\newblock Resolution adaptive networks for efficient inference.
\newblock In \emph{Proceedings of the IEEE/CVF Conference on Computer Vision
  and Pattern Recognition (CVPR)}, 2369--2378.

\bibitem[{Yu et~al.(2021)Yu, Xia, Bai, Lu, Yuille, and Shen}]{yu2021glance}
Yu, Q.; Xia, Y.; Bai, Y.; Lu, Y.; Yuille, A.~L.; and Shen, W. 2021.
\newblock Glance-and-gaze vision transformer.
\newblock In \emph{Advances in Neural Information Processing Systems
  (NeurIPS)}.

\bibitem[{Yuan et~al.(2021{\natexlab{a}})Yuan, Guo, Liu, Zhou, Yu, and
  Wu}]{yuan2021incorporating}
Yuan, K.; Guo, S.; Liu, Z.; Zhou, A.; Yu, F.; and Wu, W. 2021{\natexlab{a}}.
\newblock Incorporating convolution designs into visual transformers.
\newblock In \emph{Proceedings of the IEEE/CVF International Conference on
  Computer Vision (ICCV)}, 579--588.

\bibitem[{Yuan et~al.(2021{\natexlab{b}})Yuan, Chen, Wang, Yu, Shi, Jiang, Tay,
  Feng, and Yan}]{yuan2021tokens}
Yuan, L.; Chen, Y.; Wang, T.; Yu, W.; Shi, Y.; Jiang, Z.-H.; Tay, F.~E.; Feng,
  J.; and Yan, S. 2021{\natexlab{b}}.
\newblock Tokens-to-token vit: Training vision transformers from scratch on
  imagenet.
\newblock In \emph{Proceedings of the IEEE/CVF International Conference on
  Computer Vision (ICCV)}, 558--567.

\bibitem[{Zheng et~al.(2021)Zheng, Lu, Zhao, Zhu, Luo, Wang, Fu, Feng, Xiang,
  Torr et~al.}]{zheng2021rethinking}
Zheng, S.; Lu, J.; Zhao, H.; Zhu, X.; Luo, Z.; Wang, Y.; Fu, Y.; Feng, J.;
  Xiang, T.; Torr, P.~H.; et~al. 2021.
\newblock Rethinking semantic segmentation from a sequence-to-sequence
  perspective with transformers.
\newblock In \emph{Proceedings of the IEEE/CVF Conference on Computer Vision
  and Pattern Recognition (CVPR)}, 6881--6890.

\bibitem[{Zong et~al.(2022)Zong, Li, Song, Wang, Qiao, Leng, and
  Liu}]{zong2021self}
Zong, Z.; Li, K.; Song, G.; Wang, Y.; Qiao, Y.; Leng, B.; and Liu, Y. 2022.
\newblock Self-slimmed vision transformer.
\newblock In \emph{European Conference on Computer Vision (ECCV)}, 432--448.

\end{thebibliography}

\end{document}